\documentclass[9pt,journal]{IEEEtran}
\newlength{\wdth}
\usepackage[tight]{subfigure}
\usepackage{epsfig}
\usepackage{graphicx}
\usepackage{captcont}
\usepackage{caption}
\usepackage{float}
\usepackage{psfrag}
\usepackage{caption}
\usepackage{color}
\usepackage{amsmath,amsfonts,amssymb}
\usepackage[numbers,square, comma, sort&compress]{natbib}
\usepackage{multirow}
\date{}

\begin{document}

\title{\huge A Subband-Based SVM Front-End for Robust ASR}

\author{\normalsize{Jibran~Yousafzai$^{\dagger}$,~\IEEEmembership{\normalsize{Member,~IEEE}}~~~~Zoran~Cvetkovi\'c$^{\dagger}$,~\IEEEmembership{\normalsize{Senior~Member,~IEEE}}\\Peter~Sollich$^{\ddagger}$~\IEEEmembership{\normalsize{}}~~~~~~Matthew~Ager$^{\ddagger}$~\IEEEmembership{\normalsize{}} }

\thanks{

The authors are with the Department of Informatics$^{\dagger}$ and the Department of  Mathematics$^{\ddagger}$ at King's College London (e-mail: $\left\{\textrm{jibran.yousafzai, zoran.cvetkovic, peter.sollich, matthew.ager}\right\}$@kcl.ac.uk).}}

\newcommand{\be}{\begin{equation}}
\newcommand{\ee}{\end{equation}}

\maketitle

\begin{abstract}
\textbf{
This work proposes a novel support vector machine (SVM) based robust automatic speech recognition (ASR) front-end that operates on an ensemble of the subband components of high-dimensional acoustic waveforms. The key issues of selecting the appropriate SVM kernels for classification in frequency subbands and the combination of individual subband classifiers using ensemble methods are addressed. The proposed front-end is compared with state-of-the-art ASR front-ends in terms of robustness to additive noise and linear filtering. Experiments performed on the TIMIT phoneme classification task demonstrate the benefits of the proposed subband based SVM front-end: it outperforms the standard cepstral front-end in the presence of noise and linear filtering for signal-to-noise ratio (SNR) below 12-dB. A combination of the proposed front-end with a conventional front-end such as MFCC yields further improvements over the individual front ends across the full range of noise levels.
}
\end{abstract}



\section{INTRODUCTION}
\label{sec:sec1}

{A}UTOMATIC speech recognition (ASR) systems suffer severe performance degradation in the presence of environmental distortions, in particular additive and convolutive noise. Humans, on the other hand, exhibit a very robust behavior in recognizing speech even in extremely adverse conditions. The central premise behind the design of state-of-the-art ASR systems is that combining front-ends based on the non-linear compression of speech, such as Mel-Frequency Cepstral Coefficients (MFCC) \cite{mfcc} and Perceptual Linear Prediction (PLP) coeffcients \cite{hermansky}, with appropriate language and context modelling techniques can bring the recognition performance of ASR close to humans. However, the effectiveness of context and language modelling depends critically on the accuracy with which the underlying sequence of elementary phonetic units is predicted \cite{lippmann}, and this is where there are still significant performance gaps between humans and ASR systems. Humans recognize isolated speech units above the level of chance already at $-18$-dB SNR, and significantly above it at $-9$-dB SNR \cite{millernicely}. At such high noise levels, human speech recognition performance exceeds that of the state-of-the-art ASR systems by over an order of magnitude. Even in quiet conditions, the machine phone error rates for nonsense syllables are significantly higher than human error rates \cite{jont, lippmann,srokabraida,meyer2007}. Although there are a number of factors preventing the conventional ASR systems to reach the human benchmark, several studies \cite{atal,peters,herve,meyer2007,PaliwalAlsteris2003,PaliwalAlsteris2006} have attributed the marked difference between human and machine performance to the fundamental limitations of the ASR front-ends. These studies suggest that the large amount of redundancy in speech signals, which is removed in the process of the extraction of cepstral features such as Mel-Frequency Cepstral Coefficients (MFCC) \cite{mfcc} and Perceptual Linear Prediction (PLP) coefficients \cite{hermansky}, is in fact needed to cope with environmental distortions. Among these studies, the work on human speech perception \cite{peters,meyer2007,PaliwalAlsteris2003,PaliwalAlsteris2006} has shown explicitly that the information reduction that takes place in the conventional front-ends leads to a severe degradation in human speech recognition performance and, furthermore, that in noisy environments there is a high correlation between human and machine errors in recognition of speech with distortions introduced by typical ASR front-end processing. Over the years, techniques such as cepstral mean-and-variance normalization (CMVN) \cite{cmvn,mva}, vector Taylor series (VTS) compensation \cite{vts1} and ETSI advanced front-end (AFE) \cite{etsi} have been developed that aim to explicitly reduce the effects of noise on the short-term spectra in order to make the ASR front-ends less sensitive to noise. However, the distortion of the cepstral features caused by additive noise and linear filtering critically depends on the speech signal, filter characteristics, noise type and noise level in a complex fashion that makes effective feature compensation or adaptation very intricate and not sufficiently effective \cite{mva}.

In our previous work we showed that using acoustic waveforms directly, without any compression or nonlinear transformation
can improve the robustness of ASR front-ends to additive noise \cite{journal1}. In this paper, we propose an ASR front-end derived from the decomposition of speech into its frequency subbands, to achieve additional robustness to additive noise as well as linear filtering. This approach draws its motivation primarily from the experiments conducted by Fletcher \cite{fletcher}, which suggest that the human decoding of linguistic messages is based on decisions within narrow frequency subbands that are processed quite independently of each other. This reasoning further implies that accurate recognition in any subband should result in accurate recognition overall, regardless of the errors in other subbands. While this theory has not been proved and some studies on the subband correlation of speech signals \cite{mcauleyming2005, mingsmith2002} have even put its validity into question, there are some technical reasons for considering classification in frequency subbands. First of all, decomposing speech into its frequency subbands can be beneficial since it allows a better exploitation of the fact that certain subbands may inherently provide better separation of some phoneme classes than others. Secondly, the effect of wideband noise in sufficiently narrow subbands can be approximated as that of narrowband white noise and thus make the compensation of features  be approximately independent of the spectral characteristics of the additive noise and linear filtering. Moreover, appropriate ensemble methods for aggregation of the decisions in individual frequency subbands can facilitate selective de-emphasis of unreliable information, particularly in the presence of narrowband noise. 

Previously, the subband approach has been used in \cite{vaseghinew,sub1,sub2,sub3,sub4,sub5,sub6} which resulted in marginal improvements in recognition performance over its full band counterparts. Note that the front-ends employed in the previous works were the subband-based variants of cepstral features or multi-resolution cepstral features. By contrast, our proposed front-end features are extracted from an ensemble of subband components of high-dimensional acoustic waveforms and thus retain more information about speech that is potentially relevant to discrimination of phonetic units than the corresponding cepstral representations. In addition to investigation of robustness of the proposed front-end to additive noise, we also assess its robustness to linear filtering due to room reverberation. This form of distortion causes temporal smearing of short-term spectra which degrades the performance of ASR systems. This can be attributed primarily to the use of analysis windows for feature extraction in the conventional front-ends such as MFCC that are much shorter than typical room impulse responses. Furthermore, the distortion caused by linear filtering is correlated with the underlying speech signal. Hence, conventional methods for robust ASR that are tuned for recognition of data corrupted by additive noise only will not be effective in reverberant environments. Several speech dereverberation techniques that rely on multi-channel recordings of speech such as \cite{dereverb1,dereverb2} exist in the literature. However, these consideration extend beyond the scope of this paper and instead, standard single channel feature compensation methods for additive noise and linear filtering such as VTS and CMVN compensation are used throughout this paper.

Robustness of the proposed front-end to additive noise and linear filtering is demonstrated by its comparison with the MFCC front-end on a phoneme classification task; this task remains important in comparing different methods and representations \cite{hlmgmm,hiddencrf2,shasaul,rls2,halberstadt97,clarkson,hiddencrf,vaseghinew,galesita,classificationref1,classificationref2,classificationref3}. The improvements achieved on the classification task can be expected to extend to continuous speech recognition tasks \cite{halberstadt98,svm2} as SVMs have been employed in hybrid frameworks \cite{svm2,cont2} with hidden Markov models (HMMs) as well as in frame-based architectures using the token passing algorithm \cite{svmcontnew2} for recognition of continuous speech. The results demonstrate the benefits of the subband classification in terms of robustness to additive noise and linear filtering. The subband-waveform classifiers outperform even the MFCC classifiers trained and tested under matched conditions for signal-to-noise ratios  below 6-dB. Furthermore, in classifying noisy reverberant speech, the subband classifier outperforms the MFCC classifier compensated using VTS for all signal-to-noise ratios (SNRs) below a crossover point between 12-dB and 6-dB. Finally, their convex combination yields further performance improvements over both individual classifiers.

This paper is organized as follows: the proposed subband classification approach is described in Section \ref{sec:sec2}. Experimental results that demonstrate its robustness to additive noise and linear filtering are presented in Section \ref{sec:sec3}. Finally, Section \ref{sec:sec4} draws some conclusions and suggests future directions of this work towards application of the proposed front-end in continuous speech recognition tasks.

\section{SUBBAND CLASSIFICATION USING SUPPORT VECTOR MACHINES}
\label{sec:sec2}

Support vector machines (SVMs) are receiving increasing attention as a tool for speech recognition applications due to their good generalization properties \cite{vapnik,clarkson,galescont,svm2,cont2,sequencekernels,svm1,journal1}. Here we use them in conjunction with the proposed subband-based representation aiming to improve the robustness of
the standard cepstral front-end to noise and filtering. To this end we construct a  fixed-length representation that could potentially be used as the front-end for a continuous speech recognition systems based on \textit{e.g.} hidden Markov models (HMMs) \cite{svm2,cont2,svmcontnew2}. Dealing with variable phoneme length has been addressed by means of generative kernels such as Fisher kernels \cite{fisher,sequencekernels} and dynamic time-warping kernels \cite{svmcontnew}, but lies beyond the scope of this paper. Hence, the features of the proposed front-end are derived from fixed-length segments of acoustic waveforms of speech and these are studied in comparison with
the  MFCC features derived from the same speech segments. Several possible extensions of the proposed front-end for application in continuous speech recognition tasks are highlighted in Section \ref{sec:sec4} and will be investigated in a future study.

\subsection{Support Vector Machines}
\label{sec:sec2.1}

A binary SVM 
classifier estimates a decision surface that jointly maximizes the margin between the two classes and minimizes the misclassification error on the training set. For a given training set $\left(\mathbf{x}_1,\ldots,\mathbf{x}_p\right)$ with corresponding class labels $\left(y_1,\ldots,y_p\right),~y_i \in \{ +1,-1 \}$, an SVM classifies a test point $\mathbf{x}$ by computing a score function, 
\be h(\mathbf{x})=\sum_{i=1}^p\alpha_{i}y_{i}K(\mathbf{x},\mathbf{x}_i)+ b \ee
where $\alpha_{i}$ is the Lagrange multiplier corresponding to the $i^\textrm{th}$ training sample, $\mathbf{x}_i$, $b$ is the classifier bias -- these are optimized during training -- and $K$ is a kernel function. The class label of $\mathbf x$ is then predicted as $\mathrm{sgn}\left(h\left(\mathbf{x}\right) \right)$. While the simplest kernel $K(\mathbf{x},\tilde{\mathbf{x}})=\langle\mathbf{x},\tilde{\mathbf{x}}\rangle$ produces linear decision boundaries, in most real classification tasks, the data is not linearly separable. Nonlinear kernel functions implicitly map data points to a high-dimensional feature space where the data could potentially be linearly separable. Kernel design is therefore effectively equivalent to feature-space selection, and using an appropriate kernel for a given classification task is crucial. Commonly used is the polynomial kernel, $K_p(\mathbf{x},\tilde{\mathbf{x}})=(1 + \langle\mathbf{x},\tilde{\mathbf{x}}\rangle)^{\Theta},$ where the polynomial order $\Theta$ in $K_p$ is a hyper-parameter that is tuned to a particular classification problem. More sophisticated kernels can be obtained by various combinations of basic SVM kernels. Here we use a polynomial kernel for classification with cepstral features (MFCC) whereas classification with acoustic waveforms in frequency subbands is performed using a custom-designed kernel described in the following.

For multiclass problems, binary SVMs are combined via 
error-correcting output codes (ECOC) methods \cite{dietterichbakiri,rifkin}. In this work,  for an $M$-class problem we train $N=M(M-1)/2$ binary pairwise classifiers, primarily  to lower the computational complexity by training on only the relevant two classes of  data. The training scheme can be captured in a coding matrix $w_{mn}\in \{0,1,-1\}$, \textit{i.e.} classifier $n$ is trained only on data from the two classes $m$ for which $w_{mn}\neq 0$, with $\mathrm{sgn}(w_{mn})$ as the class label. One then predicts for test input $\mathbf{x}$ the class that minimizes the loss $\sum_{n=1}^{N}\chi(w_{mn}f_{n}(\mathbf{x}))$ where $f_{n}(\mathbf{x})$ is the output of the $n^\textrm{th}$ binary classifier and $\chi$ is a loss function. We experimented with a variety of loss functions, including hinge, Hamming, exponential and linear. The hinge loss function $\chi(z)=\max(1-z,0)$ performed best and is therefore used throughout.

For classification in frequency subbands, each waveform $\mathbf{x}$ is processed through an $S$-channel maximally-decimated perfect reconstruction cosine modulated filter bank (CMFB) \cite{martinv} and decomposed into its subband components, $\mathbf{x}^s, s=1,\ldots,S$.
Several other subband decompositions such as discrete wavelet transform, wavelet packet decomposition and discrete cosine transform also achieved comparable, albeit somewhat inferior performance. A summary of the classification results obtained with different subband decompositions in quiet conditions is presented in Section \ref{sec:sec3.2}. The CMFB consists of a set of orthonormal analysis filters
\begin{align}
g_s[k]=&\frac{1}{\sqrt{S}}g[k] \cos \left(\frac{2s-1}{4S}\left(2k-S-1\right)\pi\right), \nonumber \\
&s=1,\ldots,S,~k=1,\ldots,2S,
\label{eq:cmfb}
\end{align}
where $g[k]=\sqrt{2}\sin\left({\pi\left(k-0.5\right)}/{2S}\right),~k=1,\ldots,2S$, is a low-pass prototype filter. Such a filter bank implements an orthogonal transform, hence the collection of the subband components is a representation of the original waveform in a different coordinate system \cite{martinv}. 
The subband components $x_s[n]$ are thus given by
\begin{equation}
\mathbf{x}^s[n]=\sum_k \mathbf{x}[k]g_s[nS-k]~.
\end{equation}
A maximally-decimated filter bank was chosen primarily because the sub-sampling operation avoids introducing  additional unnecessary redundancies and thus  limits the overall computational burden. However, we believe that redundant expansions of speech signals obtained using over-sampled filter banks could be advantageous to effectively account for the shift invariance of speech.

For classification in frequency subbands, an SVM kernel is constructed by partly following steps from our previous work \cite{journal1}, which attempted to capture known invariances or express explicitly the waveform qualities which are known to correlate with phoneme identity. First, an even kernel is constructed from a baseline polynomial kernel $K_p$ to account for the sign-invariance of human speech perception as
\begin{align} \label{K_e}
K_e(\mathbf{x}^s,\mathbf{x}_i^s)&=K_p^{\prime}(\mathbf{x}^s,\mathbf{x}_i^s)+K_p^{\prime}(\mathbf{x}^s,-\mathbf{x}_i^s)~\end{align}
where $K_p^{\prime}$ is a modified polynomial kernel given by 
\begin{align}
K_p^{\prime}(\mathbf{x}^s,\mathbf{x}_i^s)&=K_p\left(\frac{\mathbf{x}^s}{\left\|\mathbf{x}^s\right\|},\frac{\mathbf{x}_i^s}{\left\|\mathbf{x}_i^s\right\|}\right)=\left(1 + \left\langle \frac{\mathbf{x}^s}{\left\|\mathbf{x}^s\right\|},\frac{\mathbf{x}_i^s}{\left\|\mathbf{x}_i^s\right\|} \right\rangle\right)^{\Theta}. \end{align}
Kernel $K_p^{\prime}$, which acts on normalized input vectors, will be used as a baseline kernel for the acoustic waveforms. On the other hand, the standard polynomial kernel $K_p$ is used for classification with the cepstral representations where feature standardization by cepstral mean-and-variance normalization (CMVN) \cite{cmvn} already ensures that feature vectors typically have unit norm. 

Next, the temporal dynamics of speech are explicitly taken into account by means of features that capture the evolution of energy in individual subbands. To obtain these features, each subband component  $\mathbf{x}^s$ is first divided into $T$ frames, $\mathbf{x}^{t,s}, t=1,\ldots,T$, and then a vector of their energies $\boldsymbol{\mathbf{\omega}}^s$ is formed as, 
$$\boldsymbol{\mathbf{\omega}}^s=\left[ \log\left\| \mathbf{x}^{1,s}\right\|^2, \ldots, \log\left\|\mathbf{x}^{T,s}\right\|^2 \right]~.$$ Finally, time differences \cite{furui} of $\boldsymbol{\mathbf{\omega}}^s$ are evaluated to form the dynamic subband feature vector $\boldsymbol{\Omega}^s$ as $\boldsymbol{\mathbf{\Omega}}^s=\left[ \boldsymbol{\mathbf{\omega}}^s ~~ \Delta \boldsymbol{\mathbf{\omega}}^s ~~ \Delta^2\boldsymbol{\mathbf{\omega}}^s \right]$. This dynamic subband feature vector $\boldsymbol{\Omega}^s$ is then combined with the corresponding acoustic waveform subband component $\mathbf{x}^s$ forming kernel $K_{\Omega}$  given by
 \be \label{K_subdelta} K_{\Omega}(\mathbf{x}^s,\mathbf{x}_i^s,\boldsymbol{\mathbf{\Omega}}^s,\boldsymbol{\mathbf{\Omega}}_i^s
)=K_e(\mathbf{x}^s,\mathbf{x}_i^s)K_p(\boldsymbol{\mathbf{\Omega}}^s,\boldsymbol{\mathbf{\Omega}}_i^s),~\ee 
where $\boldsymbol{\mathbf{\Omega}}_i^s$ is the dynamic subband feature vector corresponding to the $s^\textrm{th}$ subband component $\mathbf{x}_i^s$ of the $i$-th training point $\mathbf{x}_i$.


\subsection{Ensemble Methods}
\label{sec:sec2.4}

For each binary classification problem, decomposing an acoustic waveform into its subband components produces an ensemble of $S$ classifiers. The decision of the subband classifiers in the ensemble, given by 
\be f^s(\mathbf{x}^s,\boldsymbol{\mathbf{\Omega}}^s)=\sum_{i}\alpha_{i}^s y_{i}K_{\Omega}(\mathbf{x}^s,\mathbf{x}_i^s, \boldsymbol{\mathbf{\Omega}}^s,\boldsymbol{\mathbf{\Omega}}_i^s)+ b^s\ ,~s=1,\ldots,S \ee
are then aggregated using ensemble methods to obtain the binary classification decision for a test waveform $\mathbf{x}$. Here $\alpha_{i}^s$ and $b^s$ are the Lagrange multiplier corresponding to $\mathbf{x}_i^s$ and the bias of the $s^{\textrm{th}}$ subband binary classifier.

\subsubsection{Uniform Aggregation}
\label{sec:sec2.4.1}
Under a uniform aggregation scheme, the decisions of the subband classifiers in the ensemble are assigned uniform weights.
Majority voting is the simplest uniform aggregation scheme commonly used in machine learning. In our context it is equivalent to forming a meta-level score function as
\be h(\mathbf{x}) =  \sum_{s=1}^S \textrm{sgn}( f^s(\mathbf{x}^s,\boldsymbol{\mathbf{\Omega}}^s) ) ~, 
\label{eq:majority}
\ee
then predicting the class label as  $y=\textrm{sgn}(h(\mathbf{x}))$. In addition to this conventional majority voting scheme, which maps the  scores in individual subbands to the corresponding class labels ($\pm1$), we also considered various smooth squashing functions, \textit{e.g.} sigmoidal, as alternatives to the $\textrm{sgn}$ function in (\ref{eq:majority}), and obtained similar results. To gain some intuition about the potential of ensemble methods such as the majority voting in improving the classification performance, consider the ideal case when the errors of the individual subband classifiers in the ensemble are independent with error probability $p < 1/2$. Under these conditions, a simple combinatorial argument shows that the error probability $p_e$ of the majority voting scheme is given by 
\be p_e=\sum_{s=\lceil S/2 \rceil}^{S} {S \choose s} p^s (1-p)^{S-s}~.\ee 
where the largest contribution to the overall error is due to term with $s= \lceil S/2 \rceil$. For a large ensemble cardinality $S$, this error probability can be bounded as:
\be p_e <  p^{\lceil S/2 \rceil} (1-p)^{S-{\lceil S/2 \rceil}} \sum_{s=\lceil S/2 \rceil}^{S} {S \choose s} \approx \frac{1}{2}\left( 4p\left(1-p\right)  \right) ^{S/2}~.\ee
Therefore, in ideal conditions, the ensemble error decreases exponentially in $S$ even with this simple aggregation scheme \cite{dietensemble,hansen}. However, it has been shown that there exists a correlation between the subband components of speech and the resulting speech recognition errors in individual frequency subbands \cite{mcauleyming2005, mingsmith2002}. As a result, the majority voting scheme may not yield considerable improvements in the classification performance. Furthermore, the uniform aggregation schemes also suffer from a major drawback; they do not exploit the differences in the relative importance of individual subbands in discriminating among specific pairs of phonemes. To remedy this, we use stacked generalization \cite{wolpert} as discussed next, to explicitly learn weighting functions specific to each pair of phonemes for non-uniform aggregation of the outputs of base-level SVMs. 

\subsubsection{Stacked Generalization}
\label{sec:sec2.4.2}
Our practical implementation of stacked generalization \cite{wolpert} consists of a hierarchical two-layer SVM architecture, where the outputs of subband base-level SVMs are aggregated by a meta-level linear SVM. The decision function of the meta-level SVM classifier is of the form
\be h(\mathbf{x})=\langle\mathbf{f}(\mathbf{x}), \mathbf{w}\rangle + v = \sum_s w^s f^s(\mathbf{x}^s,\boldsymbol{\mathbf{\Omega}}^s) + v~, \ee 
where $\mathbf{f}(\mathbf{x})=\left[f^1(\mathbf{x}^1,\boldsymbol{\mathbf{\Omega}}^1), \ldots, f^S(\mathbf{x}^S,\boldsymbol{\mathbf{\Omega}}^S)\right]$ is the base-level SVM score vector of the test waveform $\mathbf{x}$, $v$ is the classifier bias, and $\mathbf{w}=\left[w^1,\ldots,w^S\right]$ is the weight vector of the meta-level classifier. Note each of the binary classifiers will have its specific weight vector  determined from an independent development/validation set $\{\tilde{\mathbf{x}}_j,\tilde{y}_j\}$. Each weight  vector can, therefore,  be expressed as
\be \mathbf{w}=\sum_j \beta_j \tilde{y}_j \mathbf{f}(\tilde{\mathbf{x}}_j), \ee 
where $\mathbf{f}(\tilde{\mathbf{x}}_j)=\left[f^1(\tilde{\mathbf{x}}^1_j,\tilde{\boldsymbol{\mathbf{\Omega}}}^1_j), \ldots, f^S(\tilde{\mathbf{x}}^S_j,\tilde{\boldsymbol{\mathbf{\Omega}}}^S_j)\right]$ is the base-level SVM score vector of the training waveform $\tilde{\mathbf{x}}_j$, and $\beta_j$ and $\tilde{y}_j$ are the Lagrange multiplier and class label corresponding to $\mathbf{f}(\tilde{\mathbf{x}}_j)$, respectively. While a base-level SVM assigns a weight to each supporting feature vector, stacked generalization effectively assigns an additional weight $w^s$ to each subband based on the performance of the corresponding base-level subband classifier. Again, ECOC methods are used to combine the meta-level binary classifiers for multiclass classification.

An obvious advantage of the subband approach for ASR is that the effect of environmental distortions in sufficiently narrow subbands can be approximated as similar to that of a narrow-band white noise. This, in turn, facilitates the compensation of features to be independent of the spectral characteristics of the additive and convolutive noise sources. In a preceding paper \cite{journal1}, we proposed an ASR front-end based on the full-band acoustic waveform representation of speech where a spectral shape adaptation of the features was performed in order to account for the varying strength of contamination of the frequency components due to the presence of colored noise. In this work, compensation of the features is performed using standard approaches such as cepstral mean-and-variance normalization (CMVN) and vector Taylor series (VTS) methods which do not require any prior knowledge of the additive and convolutive noise sources. Furthermore, we found that the stacked generalization also depends on the level of noise contaminating its training data. To this end, the weight vectors corresponding to the stacked classifiers can be tuned for classification in a particular environment by introducing similar distortion to its training data. In scenarios where a performance gain over a wide range of SNRs is desired, a multi-style training approach that offers a reasonable compromise between various test conditions can also be employed. For instance, a meta-level classifier can be trained using the score feature vectors of noisy data or the score feature vectors of a mixture of clean and noisy data. 

Note that since the dimension of the score feature vectors that form the input to the stacked subband classifier ($S$) is very small compared to the typical MFCC or waveform feature vectors, only a very limited amount of data is required to learn optimal weights of the meta-level classifiers. As such, stacked generalization offers flexibility and some coarse frequency selectivity for the individual binary classification problems, and can be particularly useful in de-emphasizing information from unreliable subbands. The experiments presented in this paper show that the subband approach attains major gains in classification performance over its full-band counterpart \cite{journal1} as well as the state-of-the-art front-ends such as MFCC.

\section{EXPERIMENTAL RESULTS}
\label{sec:sec3}

\subsection{Experimental Setup}
\label{sec:sec3.1}

Experiments are performed on the `si' (diverse) and `sx' (compact) sentences of the TIMIT database \cite{timit}. The training set consists of 3696 sentences from 168 different speakers. For testing we use the core test set which consists of 192 sentences from 24 different speakers not included in the training set. The development set consists of 1152 sentences uttered by 96 male and 48 female speakers not included in either the training or the core test set, with speakers from 8 different dialect regions. In training the meta-level subband classifiers, we use a small subset, randomly selecting an eighth of the data points in the complete TIMIT development set. The glottal stops /q/ are removed from the class labels and certain allophones are grouped into their corresponding phoneme classes using the standard Kai-Fu Lee clustering \cite{kaifulee}, resulting in a total of $M=48$ phoneme classes and $N=M(M-1)/2=1128$ classifiers. Among these classes, there are 7 groups for which the contribution of within-group confusions toward multiclass error is not counted, again following standard practice \cite{kaifulee,clarkson}. Initially, we experimented with different values of the hyperparameters for the binary SVM classifiers but decided to use fixed values for all classifiers as parameter optimization had a large computational overhead but only a small impact on the multiclass classification error: the degree of $K_p$ is set to $\Theta=6$ and the penalty parameter (for slack variables in the SVM training algorithm) to $C=1$.

\begin{figure}
\centering
{
\includegraphics[width=\wdth]{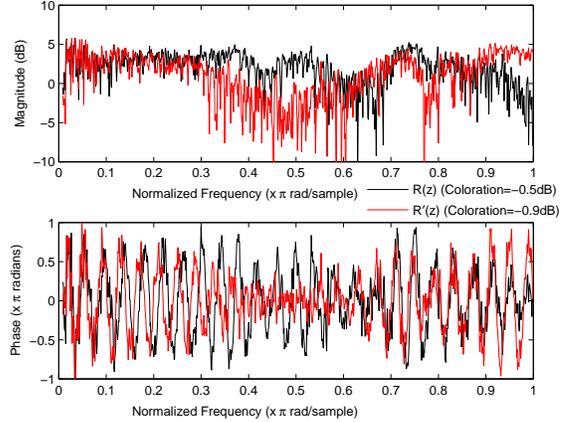}
}
\captionsetup{textfont={small,it},labelsep=colon}
\caption{Frequency response of the ICSI conference room filters with spectral coloration -0.5-dB and -0.9-dB. Here, spectral coloration is defined as the ratio of the geometric mean to the arithmetic mean of spectral magnitudes. $R(z)$ is used to add reverb to the test data whereas $R^{\prime}(z)$, a proxy filter recorded at a different location in the same room, is used for the training of cepstral and meta-level subband classifiers.}
\label{fig:fig5.4}
\end{figure}

To test the classification performance in noise, each TIMIT test sentence is normalized to unit energy per sample and then a noise sequence is added to the entire sentence to set the sentence-level SNR. Hence for a given sentence-level SNR, signal-to-noise ratio at the level of individual phonemes will vary widely. Both artificial noise (white, pink) and recordings of real noise (speech-babble) from the NOISEX-92 database are used in our experiments. White noise was selected due to its attractive theoretical interpretation as probing in an isotropic manner the separation of phoneme classes in different representation domains. Pink noise was chosen because $1/f$-like noise patterns are found in music melodies, fan and cockpit noises, in nature etc.~\cite{pink1,pink2,pink3}. In order to further test the classification performance in the presence of linear filtering, noisy TIMIT sentences are convolved with an impulse response with reverberation time $T_{60}=0.2$sec. This impulse response is one that was measured using an Earthworks QTC1 microphone in the ICSI conference room \cite{icsi} populated with people; its magnitude response $R(e^{j\omega})$ is shown in Figure \ref{fig:fig5.4}, where we also show the spectrum of an impulse response corresponding to a different speaker position in the same room, $R^{\prime}(e^{j\omega})$. While the substantial difference between these filters is evident from their spectra and spectral colorations (defined as a ratio of the geometric mean to the arithmetic mean of spectral magnitude), $R^{\prime}(e^{j\omega})$ can be viewed as an approximation of the effect of the $R(e^{j\omega})$ on the speech spectrum and is used in some of our experiments for training of the cepstral and meta-level subband classifiers in order to reduce the mismatch between training and test data.
 
To obtain the cepstral (MFCC) representation, each sentence is converted into a sequence of 13 dimensional feature vectors, their time derivatives and second order derivatives which are combined into a sequence of 39 dimensional feature vectors. Then, $T=10$ frames (with frame duration of $25$ms and a frame rate of $100$ frames/sec) closest to the center of a phoneme are concatenated to give a representation in $\mathbb{R}^{390}$. Noise compensation of the MFCC features is performed via vector Taylor series (VTS) method which has been extensively used in recent literature and is considered as state-of-the-art. This scheme estimates the distribution of noisy speech given the distribution of clean speech, a segment of noisy speech, and the Taylor series expansion that relates the noisy speech features to the clean ones, and then uses it to predict the unobserved clean cepstral feature vectors. In our experiments, a Gaussian mixture model (GMM) with 64 mixture components was used to learn the distribution of the Mel-log spectra of clean training data. Additionally, cepstral mean-and-variance normalization (CMVN) \cite{cmvn,mva} is performed to standardize the cepstral features, fixing their range of variation for both training and test data. CMVN computes the mean and variance of the feature vectors across a sentence and standardizes the features so that each has zero mean and a fixed variance. The following training-test scenarios are considered for classifiers with the cepstral front-end: 
\begin{itemize}
\item[1.] \textbf{Anechoic training with VTS} - training of the SVM classifiers is performed with anechoic clean speech and the test data is compensated via VTS.
\item[2.] \textbf{Reverberant training with VTS} - training of the SVM classifiers is performed with reverberant clean speech with feature compensation of the test data via VTS. Two particular cases in this scenario are considered. \textit{(a)} The clean training data and the noisy test data are convolved using the same linear filter, $R(e^{j\omega})$). This case provides a lower bound on the classification error in the unlikely event when the exact knowledge of the convolutive noise source is known. \textit{(b)} This case investigates the effects of a mismatch of the linear filter used for convolution with the training and test data. In particular, the data used for training of the SVM classifiers as well as learning of the distribution of log-spectra in VTS feature compensation is convolved with $R^{\prime}(e^{j\omega})$ while the test data is convolved with $R(e^{j\omega})$). Since the exact knowledge of the linear filter corrupting the test data is usually difficult to determine, this case offers a more practical solution to the problem and its performance is expected to lie between the brackets obtained with the two scenarios mentioned above \textit{i.e.} anechoic training, and reverberant training and testing using the same filter.
\item[3.] \textbf{Matched training} - In this scenario, both the training and testing conditions are identical. Again, this is an impractical target; nevertheless, we present the results (only in the presence of additive noise) as a reference, since this setup is considered to give the optimal achievable performance with cepstral features \cite{multistyle1,multistyle2,gales96robust}.
\end{itemize}
Furthermore, note that the MFCC features of both training and test data are standardized using CMVN \cite{cmvn} in all scenarios. 

Acoustic waveforms segments $\mathbf{x}$ are extracted from the TIMIT sentences by applying a 100ms rectangular window at the centre of each phoneme and are then decomposed into subband components $\{\mathbf{x}^s\}_{s=1}^{S}$ using a cosine-modulated filter bank (see \eqref{eq:cmfb}). We conducted experiments to examine the effect of the number of filter bank channels ($S$) on classification accuracy. Generally, decomposition of speech into wider subbands does not effectively capture the frequency-specific dynamics of speech and thus results in relatively poor performance. On the other hand, decomposition of speech in sufficiently narrow subbands improves classification performance as demonstrated in \cite{sub1}, but at the cost of an increase in the overall computational complexity. For the results presented in this paper, the number of filter bank channels is limited to $16$ in order to reduce the computational complexity. The dynamic subband feature vector, $\boldsymbol{\Omega}^s$ is computed by extracting $T=10$ equal-length (25ms with an overlap of 10ms) frames around the centre of each phoneme thus yielding a vector of dimension $30$. These feature vectors are further standardized within each sentence of TIMIT for the evaluation of kernel $K_{\Omega}$. Note that the training of base-level SVM subband classifiers is always performed with clean data. The development subset is used for training of the meta-level subband classifiers as learning the optimal weights requires only a limited amount of data. Several scenarios are considered for training of the meta-level classifiers:
\begin{itemize}
\item[1.] \textbf{Anechoic clean training} - training the meta-level SVM classifier with the base-level SVM score vectors obtained from anechoic clean data.
\item[2.] \textbf{Anechoic multi-style training} - training the meta-level SVM classifier with the base-level SVM score vectors of anechoic data containing a mixture of clean waveforms and waveforms corrupted by white noise at 0-dB SNR,
\item[3.] \textbf{Reverberant multi-style training} - training the meta-level SVM classifier with the base-level SVM score vectors of reverberant data containing a mixture of clean waveforms and waveforms corrupted by white noise at 0-dB SNR. Similar to the MFCC training-test scenarios, two particular cases are considered here: \textit{(a)} the development data for training as well as the test data are convolved with the same filter $R(e^{j\omega})$, and \textit{(b)} the development data for training is convolved with $R^{\prime}(e^{j\omega})$ whereas the test data is convolved with $R(e^{j\omega})$,
\item[4.] \textbf{Matched training} - training and testing with the meta-level classifier under identical noise level and type conditions. Results for this scenario are shown only in the presence of additive noise.
\end{itemize}
Next, we present the results of TIMIT phoneme classification with the setup detailed above.

\subsection{Results: Robustness to Additive Noise}
\label{sec:sec3.2}

First we compare various frequency decompositions and ensemble methods for subband classification. A summary of their respective classification errors in quiet condition is presented in Table \ref{table5.1}. We find the stacked generalization to yield significantly better results than majority voting; it consistently achieves over $10\%$ improvement over majority voting for all subband decompositions considered here. Among these decompositions, classification with the 16-channel cosine-modulated filter bank achieves the largest improvement of $5.5\%$ over the composite acoustic waveforms \cite{journal1} and is therefore selected for further experiments.

\begin{table}[th]
\captionsetup{textfont={small,it}}
\caption{Errors obtained with different subband decompositions \cite{martinv} (listed in the left column) and aggregation schemes for subband classification in quiet condition.}
\centering
\begin{tabular}{|l||c|c|}
\hline
& \multicolumn{2}{|c|}{\textbf{ERROR [$\%$]}} \\ \hline
\textbf{Subband Analysis} & \textbf{Maj. Voting} & \textbf{Stack. Gen.} \\ \hline
Level-4 wavelet decomposition & $43.7$ & $\mathbf{31.8}$ \\ \hline
Level-4 wavelet packet decomposition & $45.1$ & $\mathbf{33.1}$ \\ \hline
DCT (16 uniform-width bands) & $44$ & $\mathbf{32.6}$ \\ \hline
16-channel CMFB & $42.4$ & $\mathbf{31.2}$ \\ \hline
\textbf{Composite Waveform} \cite{journal1} & \multicolumn{2}{|c|}{$\mathbf{36.7}$} \\ \hline
\end{tabular}
\label{table5.1}
\end{table}
\vspace{8mm}

\begin{figure}
\centering
\includegraphics[width=\wdth]{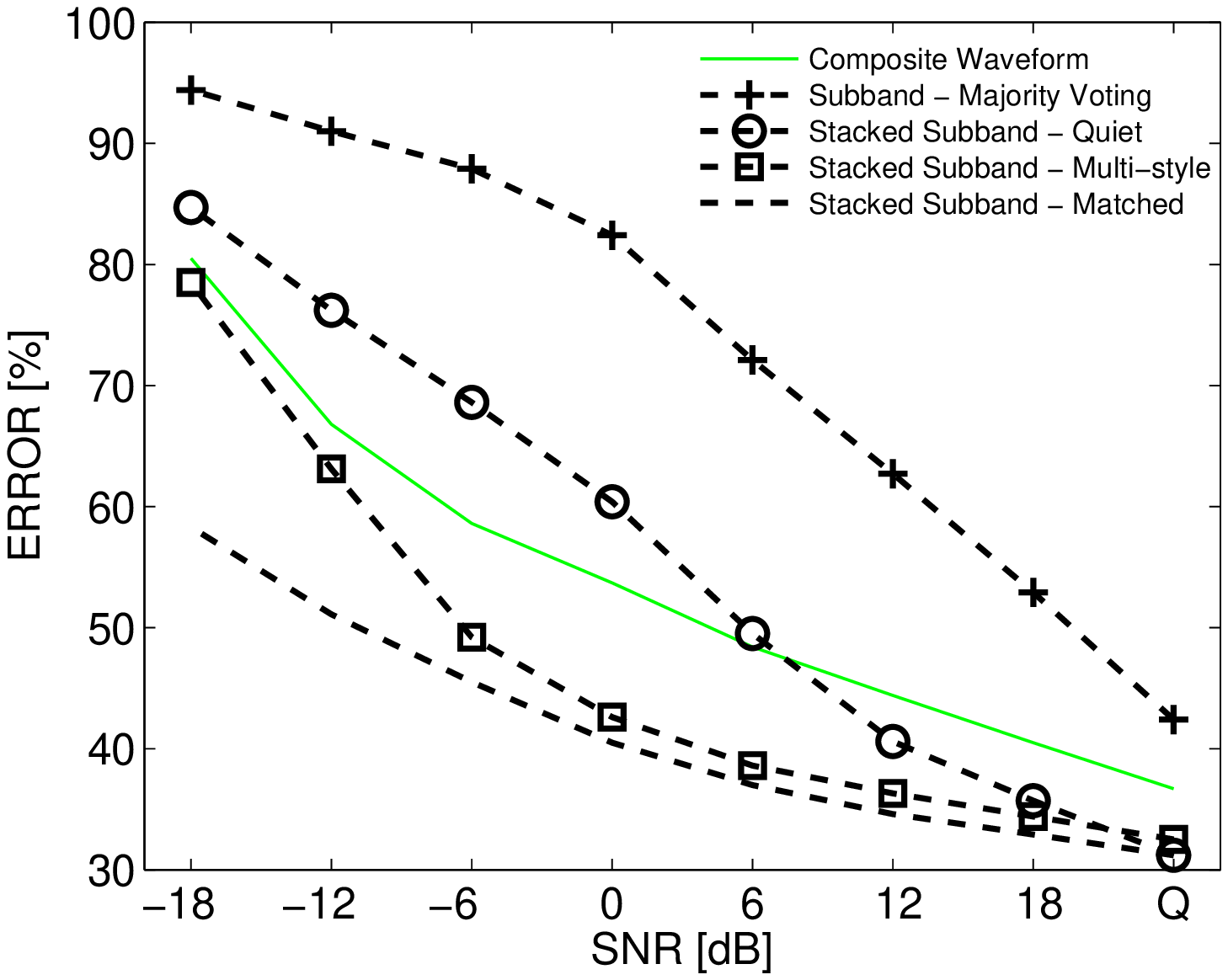}
\includegraphics[width=\wdth]{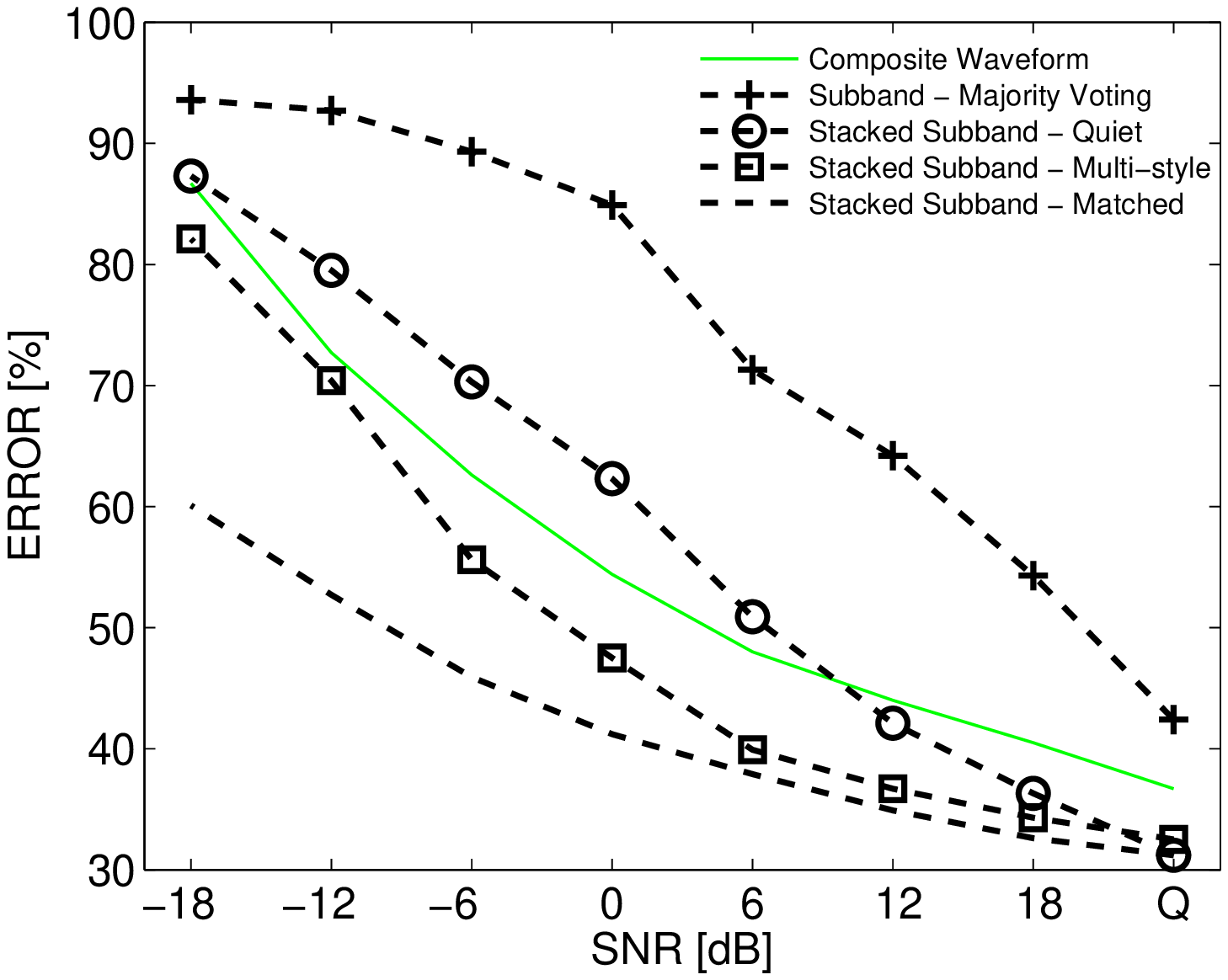}
\captionsetup{textfont={small,it}}
\caption{Ensemble methods for aggregation of subband classifiers and their comparison with composite acoustic waveform classifiers (results as reported in \cite{journal1}) in the presence of white noise (top) and pink noise (bottom). The curves correspond to uniform combination (majority voting) and stacked generalization with different training scenarios for the meta-level classifiers. The multi-style stacked subband classifier is trained only with the small development subset (one eighth randomly selected score vectors from the development set) consisting of clean and white-noise (0-dB SNR) corrupted anechoic data. The classifiers are then tested on data corrupted with white noise (matched) and pink noise (mismatched).}
\label{fig:fig5.2}
\end{figure}

\begin{figure}
\centering
{
\includegraphics[width=\wdth]{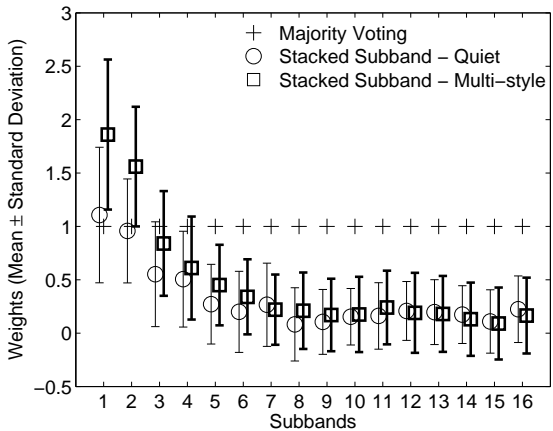}
}
\captionsetup{textfont={small,it}}
\caption[Weights (mean $\pm$ standard deviation across binary classifiers) assigned to the 16 subbands by the multi-style meta-level classifier and by the meta-level classifier trained in quiet conditions]{Weights (mean $\pm$ standard deviation across $N=1128$ binary classifiers) assigned to $S=16$ subbands by the multi-style meta-level classifiers and by the meta-level classifiers trained in quiet conditions.}
\label{fig:fig5.2.3}
\end{figure}

Let us now consider classification of phonemes in the presence of additive noise. Robustness of the proposed method to both additive noise and linear filtering is discussed in Section \ref{sec:sec3.3}. In Figure \ref{fig:fig5.2}, we compare the classification in frequency subbands using ensemble methods with composite acoustic waveform classification (results as reported in \cite{journal1}) in the presence of white and pink noise. The dashed curves correspond to subband classification using ensemble methods \textit{i.e} uniform combination (majority voting) and stacked generalization with different training scenarios for meta-level classifiers (see Section \ref{sec:sec3.1}). The meta-level classifiers of multi-style stacked subband classifier are trained according to scenario 2. The results show that stacked generalization generally attains better performance than uniform aggregation. The majority voting scheme also performs poorly in comparison to the composite acoustic waveforms across all SNRs. On the other hand the stacked subband classifier trained  in quiet condition improves over the composite waveform classifier in low noise conditions. But its performance then degrades relatively quickly in high noise because its corresponding meta-level binary classifiers are trained to assign weights to different subbands that are tuned for classification in quiet. To improve the robustness to additive noise, the meta-level classifiers can be trained on a mixture of base-level SVM score vectors obtained from both clean and data corrupted by white noise (0-dB SNR), as explained above. Figure \ref{fig:fig5.2.3} shows the weights (mean $\pm$ standard deviation across $N=1128$ binary classifiers) assigned to $S=16$ subbands by the stacked classifier with its metal-level binary classifiers trained in quiet and in multi-style conditions. It can be observed that relatively high weights are assigned to the low frequency subband components by the multi-style training. This is reasonable as low frequency subbands hold a substantial portion of speech energy and can provide reliable discriminatory information in the presence of wideband noise. The large amount of variation in the assigned weights as indicated by the error bars is consistent with the variation of speech data encountered by the $N=1128$ binary phoneme classifiers. It can be observed that the multi-style subband classifiers consistently improves over the composite waveform classifier as well as the stacked subband classifier trained in quiet condition. Overall, it achieves average improvements of $6.8\%$ and $5.9\%$ over the composite waveform classifier in the presence of white (matched noise type) and pink (mismatched noise type) noise, respectively. As expected, the stacked subband classifier trained in matched conditions finally outperforms the other classifiers in all noise conditions as shown in Figure \ref{fig:fig5.2}.

\begin{figure}
\centering
\includegraphics[width=\wdth]{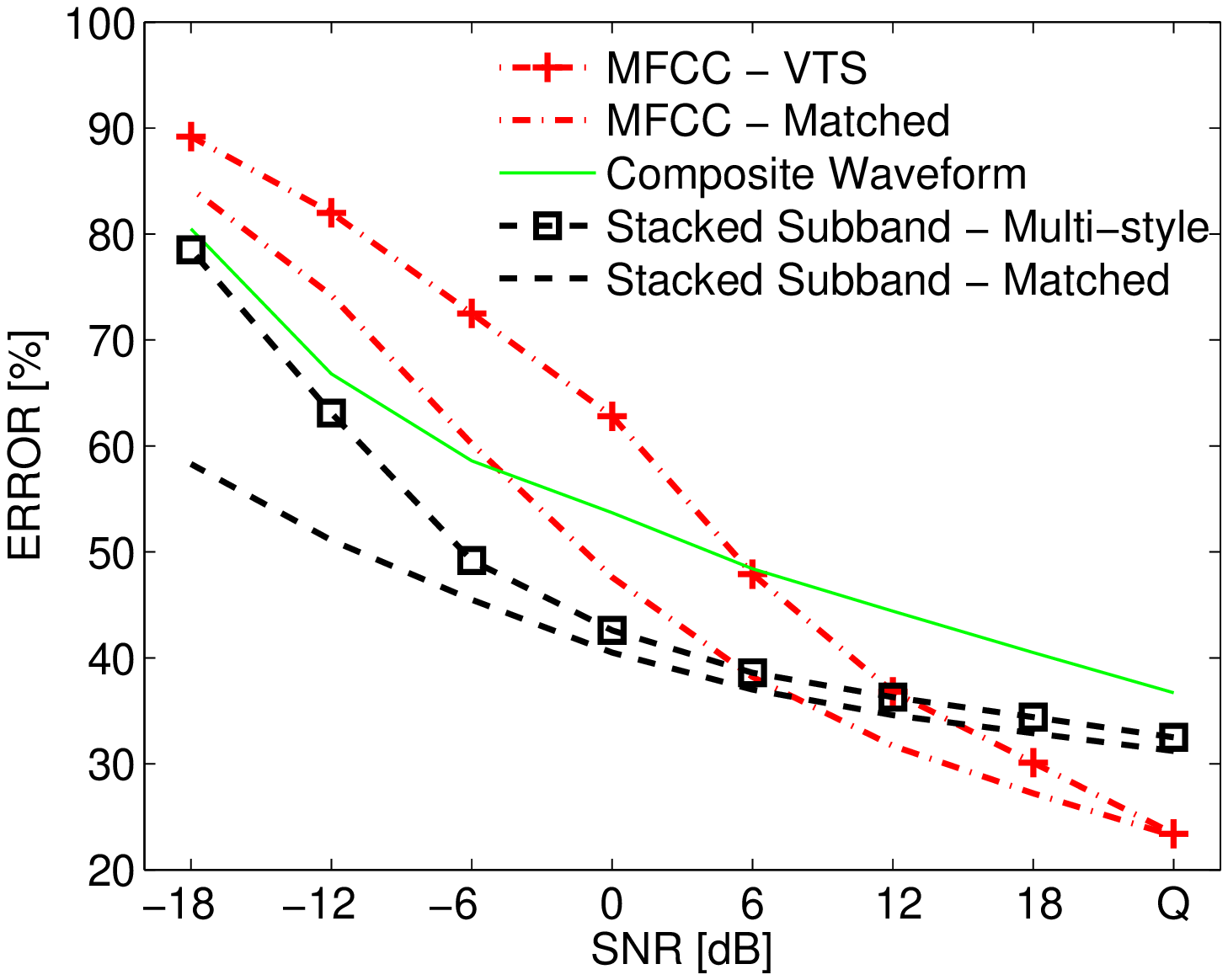}
\includegraphics[width=\wdth]{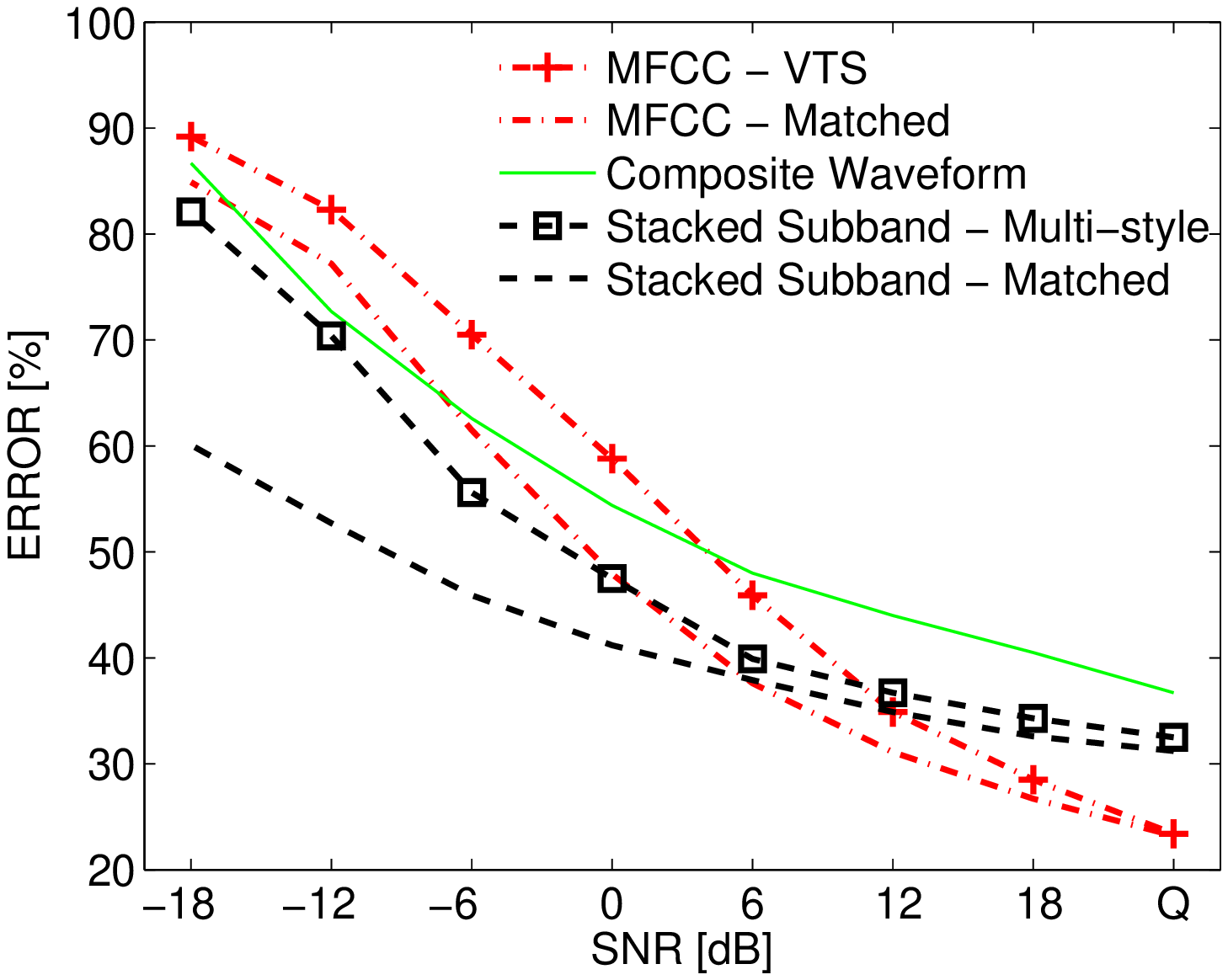}
\captionsetup{textfont={small,it},labelsep=colon}
\caption{SVM classification in the subbands of acoustic waveforms and its comparison with MFCC and composite acoustic waveform classifiers in the presence of white noise (top) and pink noise (bottom). The multi-style stacked subband classifier is trained only with a small subset of the development data (one eighth randomly selected score vectors from the development set) consisting of clean and white-noise (0-dB SNR) corrupted data. In the matched training case, noise levels as well as noise types of training and test data are identical for both MFCC and stacked subband classifiers.}
\label{fig:fig5.3}
\end{figure}

Next, we compare the performance of the multi-style subband classifier with the VTS-compensated MFCC classifier and the composite acoustic waveform classifier \cite{journal1} in the presence of additive white and pink noise. These results along with classification with the stacked subband classifier and MFCC classifier in matched training-test conditions are presented in Figure \ref{fig:fig5.2}. The results show that the stacked subband classifier exhibits better classification performance than the VTS-compensated MFCC classifier for SNR below 12-dB whereas the performance crossover between MFCC and composite acoustic waveform classifiers is between 6-dB and 0-dB SNR. The stacked subband classifier achieves average improvements of $8.7\%$ and $4.5\%$ over the MFCC classifier across the range of SNRs considered in the presence of white and pink noise, respectively. Moreover, and quite remarkably, the stacked subband classifier also significantly improves over the MFCC classifier trained and tested in matched conditions for SNRs below a crossover point between 6-dB and 0-dB SNR, even though its meta-level classifiers are trained only using clean data and data corrupted by white noise at 0-dB SNR and the number of data points used to learn the optimal weights amounts only to a small fraction of the data set used for training of the MFCC classifier in matched conditions. In particular, an average improvement of $6.5\%$ in the phoneme classification error is achieved by the multi-style subband classifier over the matched MFCC classifier for SNRs below 6-dB in the presence of white noise. 

In \cite{isit} we showed that the MFCC classifiers suffer performance degradation in case of a mismatch of the noise type between training and test data. On the other hand, the stacked subband classifier degrades gracefully in a mismatched environment as shown in Figure \ref{fig:fig5.3}. This can be attributed to the decomposition of acoustic waveforms into frequency subbands where the effect of wideband colored noise on each binary subband classifier can be approximated as that of a narrow-band white noise. In comparison to the result reported in \cite{rls2}, where a $77.8\%$ error was obtained at $0$-dB SNR in pink noise using a second-order regularized least squares algorithm (RLS2) trained using MFCC feature vectors with variable length encoding, the proposed method achieves a $30\%$ relative improvement with a fixed length representation in similar conditions. 

Figure \ref{fig:fig5.3} also shows a comparison of the stacked subband classifier with the MFCC classifier trained and tested in matched conditions. The matched-condition subband classifier significantly outperforms the matched MFCC classifier for SNRs below 6-dB. Around $13\%$ average improvement is achieved by the subband classifier over MFCC classifier for SNRs below 6-dB in the presence of both white and pink noise. This suggests that  the high-dimensional subband representation obtained from acoustic waveforms might provide a better separation of phoneme classes compared to cepstral representation. 

\subsection{Results: Robustness to Linear Filtering}
\label{sec:sec3.3}

\begin{figure}
\centering
\includegraphics[width=\wdth]{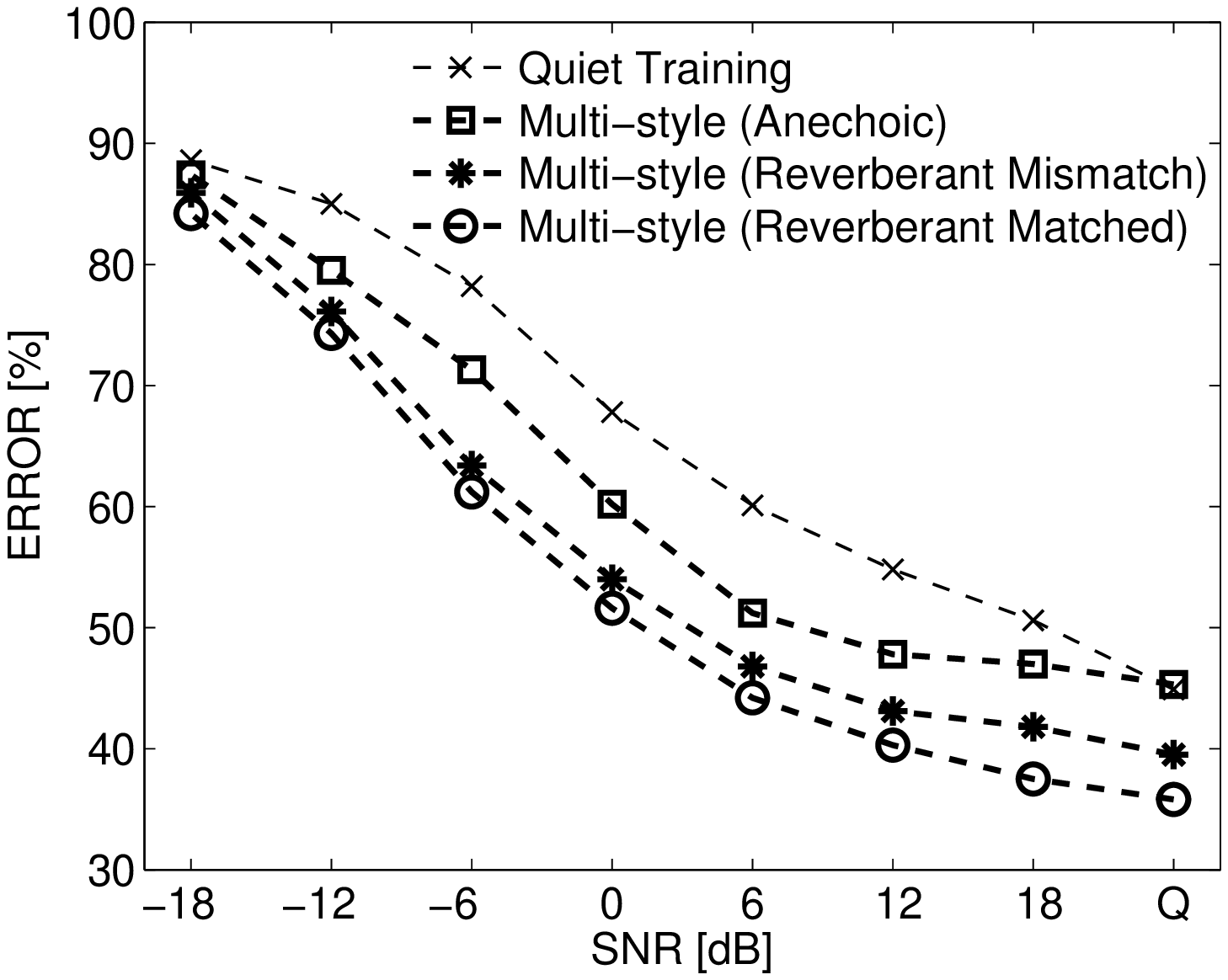}
\includegraphics[width=\wdth]{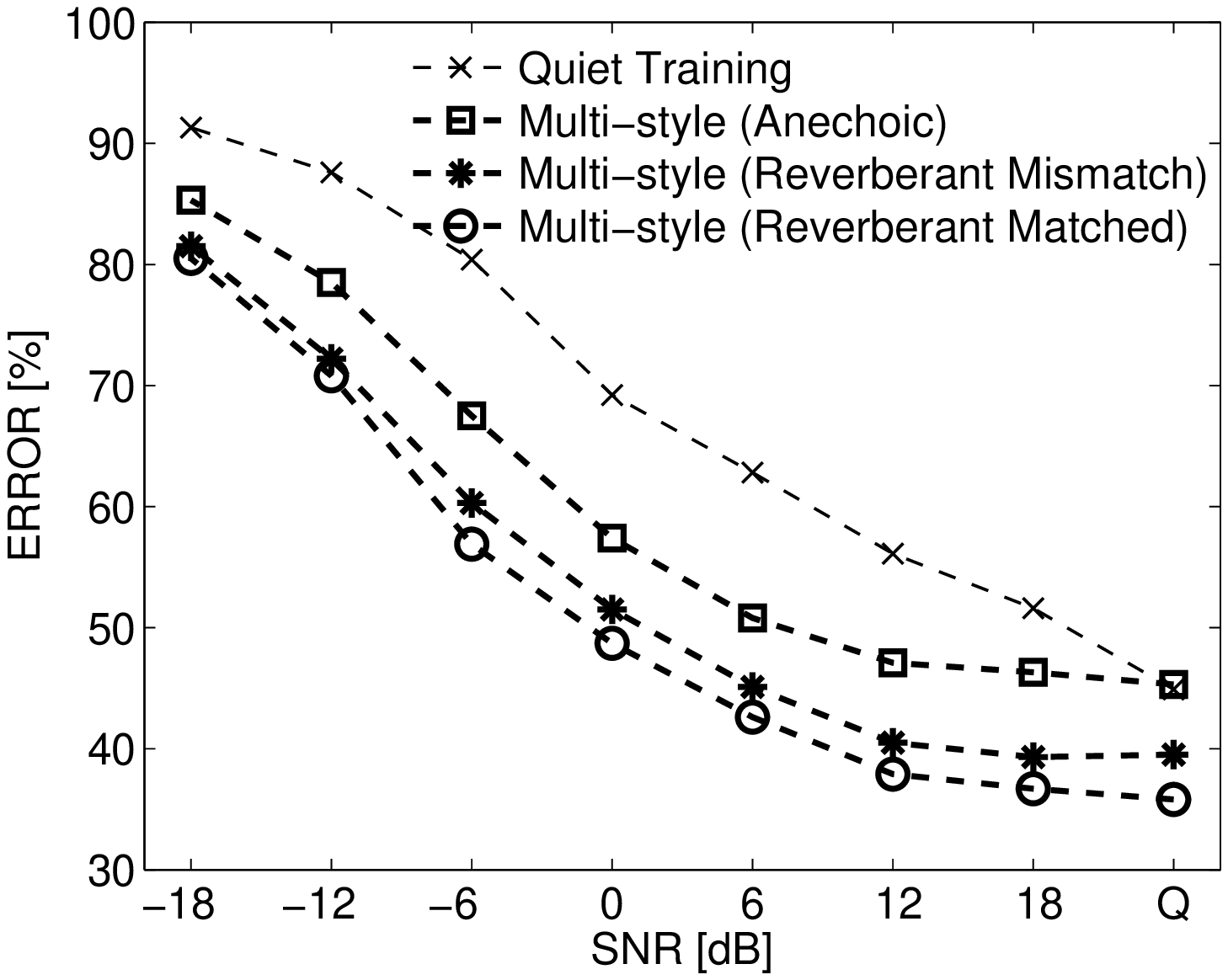}
\captionsetup{textfont={small,it},labelsep=colon}
\caption{Classification in frequency subbands using ensemble methods in the presence of the linear filter $R(e^{j\omega})$ with white noise (top) and pink noise (bottom). The curves correspond to stacked generalization with different training scenarios for meta-level subband classifier.}
\label{fig:fig5.6}
\end{figure}

We now consider classification in the presence of additive noise as well as linear filtering. First, Figure \ref{fig:fig5.6} presents results of the ensemble subband classification using stacked generalization with multiple training-test scenarios (see Section \ref{sec:sec3.1}) in the presence of white and pink noise. To reiterate, three different scenarios are considered for training of the multi-style stacked subband classifier: one involves training the meta-level classifiers with the base-level SVM score vectors of the development subset consisting of clean and white-noise (0-dB SNR) corrupted anechoic data, second involves training with the score vectors of the same development data convolved with $R^{\prime}(e^{j\omega})$ (mismatched reverberant conditions) while the third involves training in matched reverberant conditions \textit{i.e.} training with the same development subset convolved with $R(e^{j\omega})$. These classifiers, which are referred to as \textit{anechoic} and \textit{reverberant} multi-style subband classifiers (see Section \ref{sec:sec3.1}), are then tested on data corrupted by white, pink or speech-babble noise, and convolved with $R(e^{j\omega})$. Similar to our findings in the previous section, the results in Figure \ref{fig:fig5.6} show that the anechoic multi-style subband classifier consistently improves over the stacked subband classifier trained only in quiet condition. Moreover, the reverberant multi-style subband classifiers (both matched and mismatched) further reduce the mismatch with the test data and hence exhibit more superior performances. For instance, in the presence of pink noise and linear filtering, the subband classifiers trained in mismatched and matched reverberant conditions attain average improvements of $6\%$ and $8.5\%$ across all SNRs over the anechoic multi-style subband classifier, respectively. Note that an accurate measurement of the linear filter corrupting the test data may be difficult to obtain in practical scenarios. Nonetheless, classification results in matched reverberant condition are presented to determine a lower bound on the error. On the other hand, the mismatched reverberant case can be considered as a more practical solution to the problem and its performance is expected to lie between the brackets obtained with the anechoic training and matched reverberant training.

\begin{figure}
\centering
\includegraphics[width=\wdth]{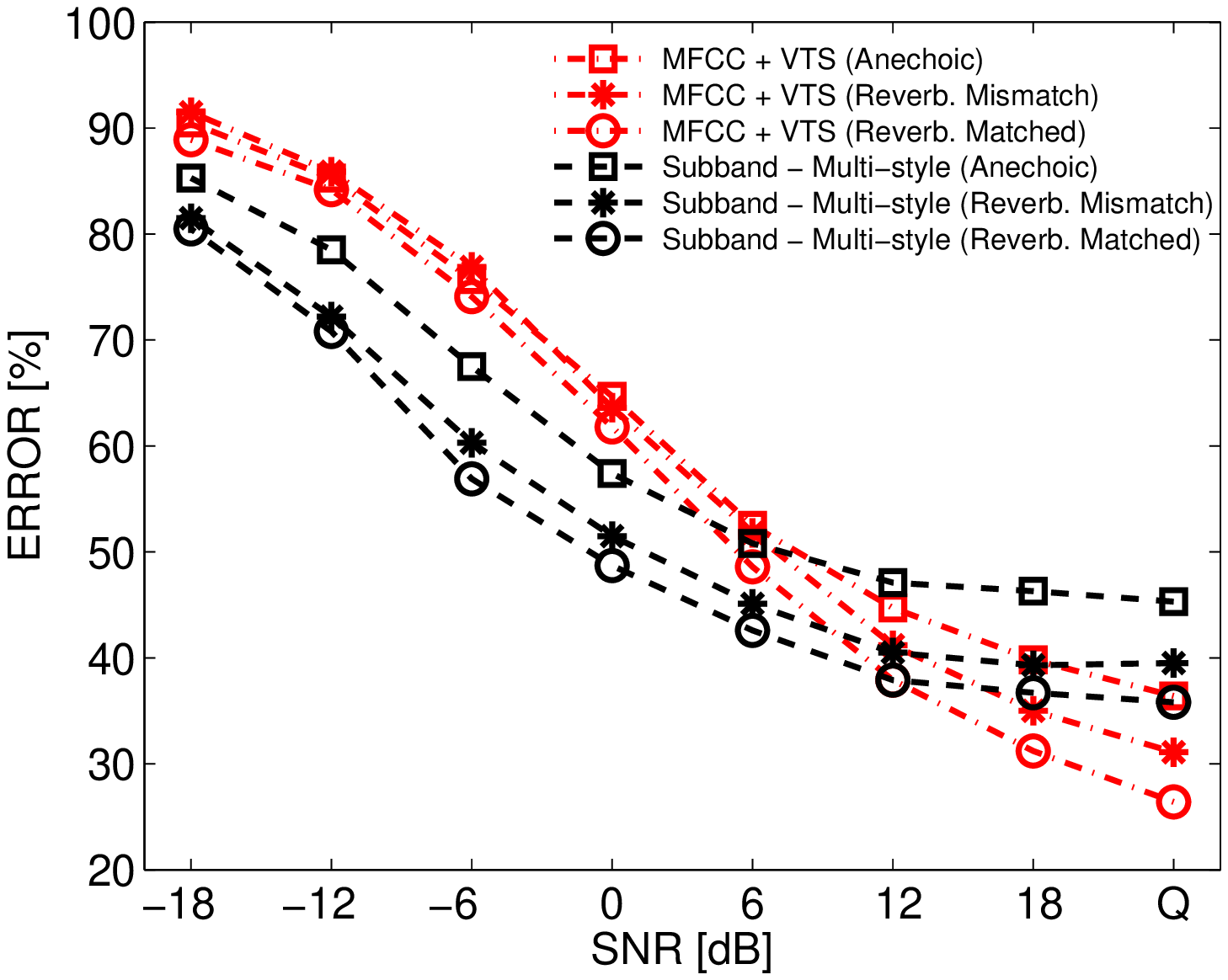}
\includegraphics[width=\wdth]{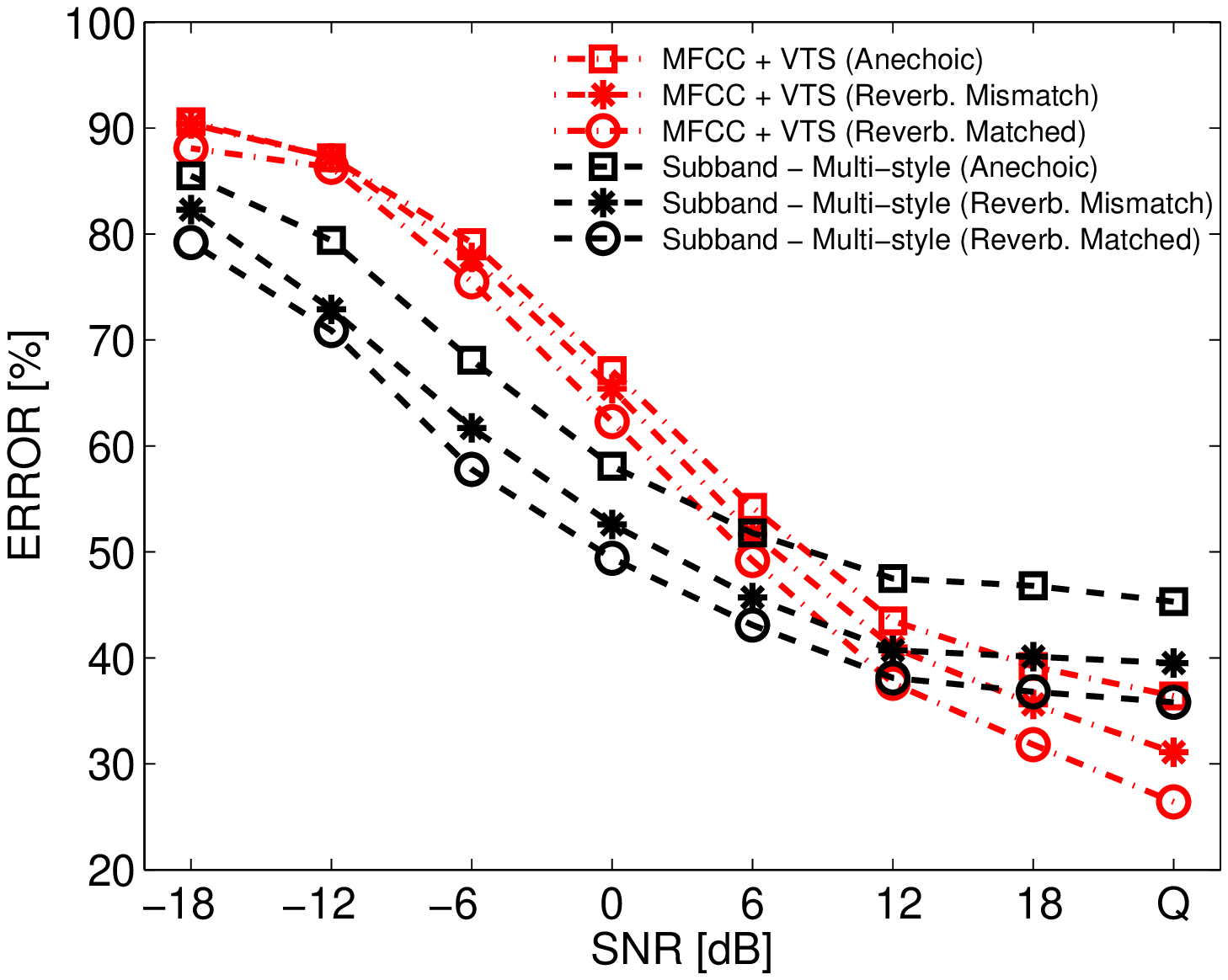}
\captionsetup{textfont={small,it},labelsep=colon}
\caption{Classification with the subband and VTS-compensated MFCC classifiers trained under three different scenarios: anechoic training (squares), reverberant mismatched training (stars) and reverberant matched training (circles). Classification results for the test data contaminated with pink noise (top) and speech-babble noise (bottom), and linear filter $R(e^{j\omega})$ are shown.}
\label{fig:fig5.7}
\end{figure}

Figure \ref{fig:fig5.7} compares the classification performances of the subband and VTS-compensated MFCC classifiers trained under three different scenarios (see Section \ref{sec:sec3.1}) in the presence of linear filtering, and pink and speech-babble noise. The first is an agnostic (anechoic) case that does not rely on any information at all regarding the source of the convolutive noise $R(e^{j\omega})$, the second (reverberant mismatch case) employs a proxy reverberation filter $R^{\prime}(e^{j\omega})$ in order to reduce the mismatch of the training and the reverberant test environments up to a certain degree, whereas the third (reverberant matched case) employs accurate knowledge of the reverberation filter $R(e^{j\omega})$ in the training of the MFCC classifiers and the meta-level subband classifiers. These training scenarios are respectively represented by squares, stars and circles in Figure \ref{fig:fig5.7}. The results show that the comparisons of the stacked subband classifiers and MFCC classifiers under the different training regimes exhibit similar trends. Generally speaking, the MFCC classifier outperforms the corresponding subband classifier in quiet and low noise conditions however the latter yield significant improvements in high noise conditions. For example, the anechoic subband classifiers yields better classification performance than the anechoic MFCC classifier for SNRs below a crossover point between $12$-dB and $6$-dB. Quantitatively similar conclusions apply to the comparative performances of the MFCC and subband classifiers in the reverberant training scenarios. Under the three different training regimes and two different noise types, the subband classifiers attain an average improvement of $8.2\%$ over the MFCC classifiers across all SNRs below $12$-dB. Note that in the reverberant training scenarios, the MFCC classifier is trained with the complete TIMIT reverberant training set. On the other hand, the meta-level subband classifier is trained using the reverberant development subset with a number of data points less than $4\%$ of that in the TIMIT training set. Moreover, the dimension of the feature vectors that form the input to the meta-level classifiers is almost 24 times smaller than the MFCC feature vectors. To this end, the subband approach offers more flexibility in terms of training and adaptation of the classifiers to a new environment.

Since an obvious performance crossover between the subband and MFCC classifiers exists at moderate SNRs, we therefore consider a convex combination of the scores of the SVM classifiers with a combination parameter $\lambda$ as discussed in \cite{journal1}. Here $\lambda=0$ corresponds to the MFCC classification whereas $\lambda=1$ corresponds to the subband classification. The combination approach was also motivated by the differences in the confusion matrices of the two classifiers (not shown here). This suggests that the errors of the subband and MFCC classifiers may be independent up to a certain degree and therefore a combination of the two may yield better performance than either of classifiers individually. Two different values of the combination parameter $\lambda$ are considered. First, the value of $\lambda$ is set to $1/2$ which corresponds to the arithmetic mean of the MFCC and subband SVM classifier scores. In the second case, we set the combination parameter $\lambda$ to a function $\lambda_{\textrm{emp}}(\sigma^2)$ which approximates the optimal combination parameter values for an independent development set. This approximated function was determined empirically in our previous experiments \cite{journal1} and is given by $\lambda_{\textrm{emp}}(\sigma^2) = \eta + {\zeta}/[{1+\left(\sigma_0^2/\sigma^2\right)}]$,
with $\eta=0.2$, $\zeta=0.5$  and $\sigma_0^2=0.03$. Note that $\lambda_{\textrm{emp}}(\sigma^2)$ also requires an estimate of the noise variance $(\sigma^2)$ which was explicitly measured using the decision-directed estimation algorithm \cite{malah,malah2}.

\begin{figure}
\centering
\includegraphics[width=\wdth]{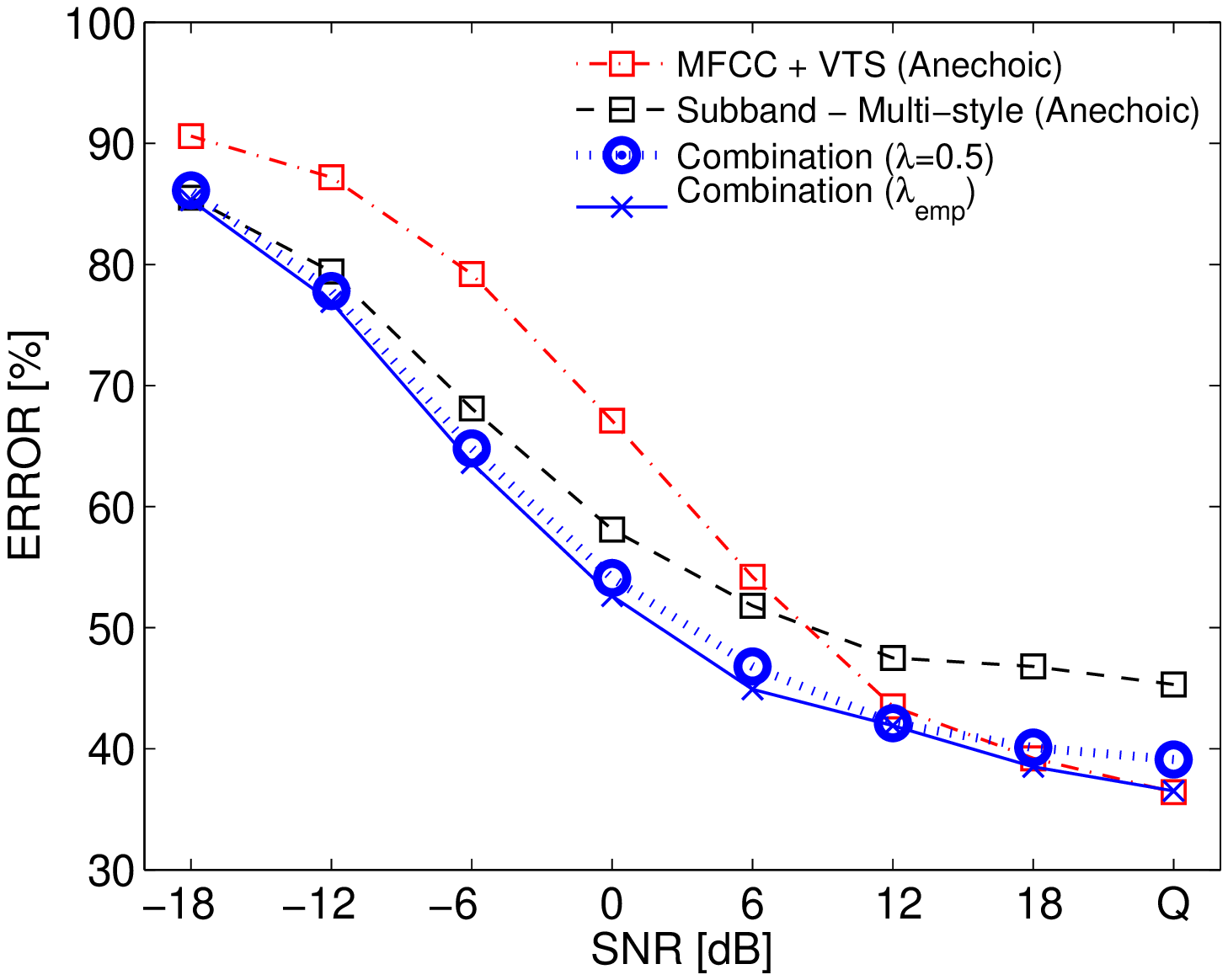}
\includegraphics[width=\wdth]{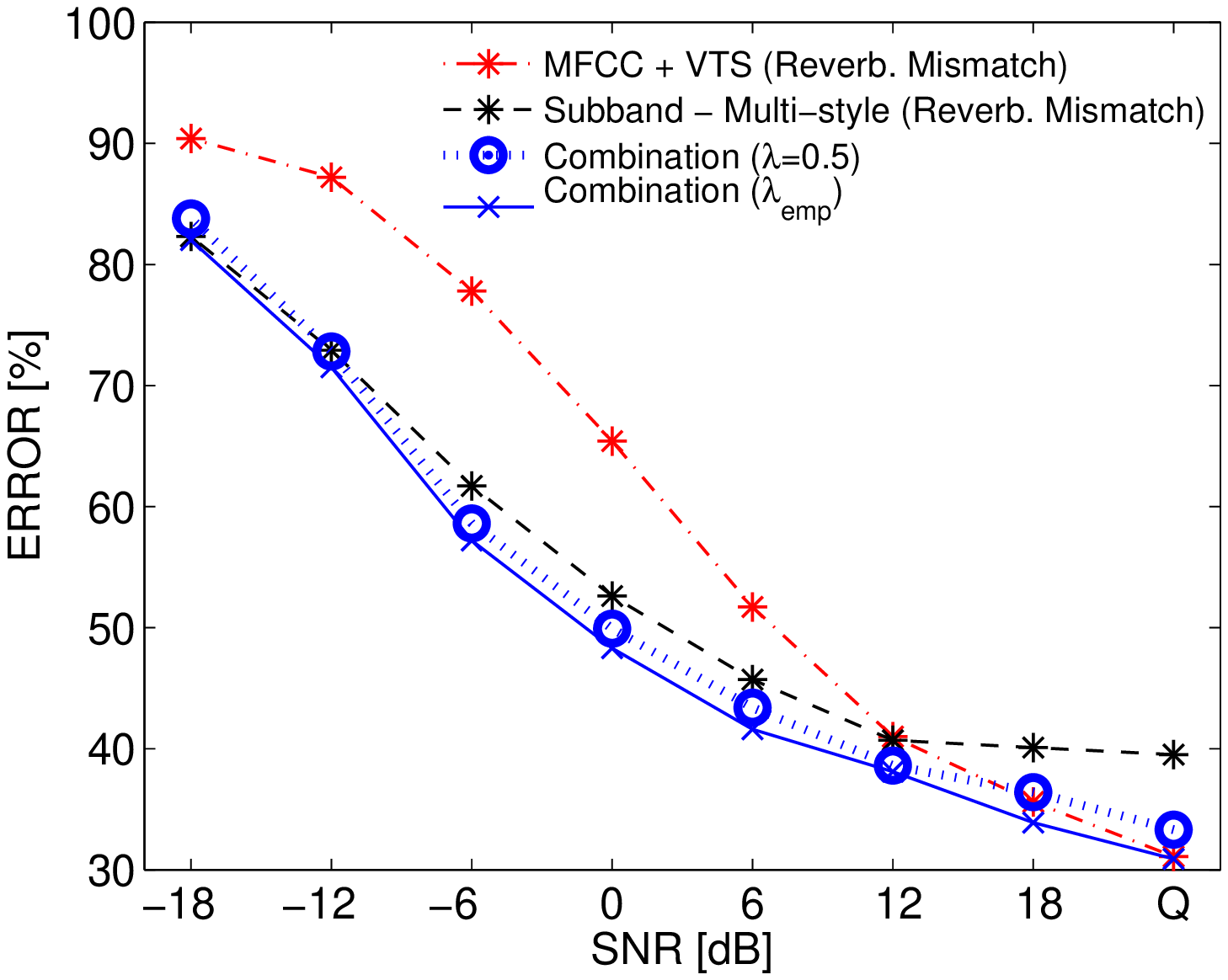}
\captionsetup{textfont={small,it},labelsep=colon}
\caption{Comparison of the classification performances of the subband and MFCC classifiers with their convex combination in the presence of speech-babble noise under anechoic training (top) and reverberant mismatched training regimes (bottom). Results are shown for two different settings of the combination parameter, $\lambda$.}
\label{fig:fig5.8}
\end{figure}

Figure \ref{fig:fig5.8} compares the classification performances of the subband and MFCC classifiers with their convex combination in the presence of speech-babble noise under anechoic and reverberant mismatched training regimes. One can observe that the combined classification with $\lambda_{\textrm{emp}}$ consistently outperforms either of the individual classifiers across all SNRs. For instance, under the anechoic training of the classifiers, the combined classification with $\lambda_{\textrm{emp}}$ attains a $5.3\%$ and $7.2\%$ average improvement over the subband and MFCC classifiers respectively, across all SNRs considered. Moreover, the combined classification via a simple averaging of the subband and MFCC classifiers by setting $\lambda=1/2$ provides a reasonable compromise between classification performance achieved within both representation domains i.e. subbands of acoustic waveforms and cepstral representation. While the performance of the combined classifier with $\lambda=1/2$ degrades only slightly (approximately $2\%$) for SNRs above a cross over point between $18$-dB and $12$-dB, it achieves relatively far greater improvements in high noise. e.g. under the anechoic training regime, the combined classifier with $\lambda=1/2$ attains a $13\%$ and $4.2\%$ improvement over the MFCC and subband classifiers at $0$-dB SNR, respectively. Quantitatively similar conclusions apply in the reverberant mismatched training scenario as shown in Figure \ref{fig:fig5.8}.

\section{CONCLUSIONS}
\label{sec:sec4}

In this paper we studied an SVM front-end for robust speech recognition that operates in frequency subbands of high-dimensional acoustic waveforms. We addressed the issues of kernel design for subband components of acoustic waveforms and the aggregation of the individual subband classifiers using ensemble methods. The experiments demonstrated that the subband classifiers outperform the cepstral classifiers in the presence of noise and linear filtering for SNRs below $12$-dB. While the subband classifiers do not perform as well as the MFCC classifiers in low noise conditions, major gains across all noise levels can be attained by a convex combination \cite{journal1}. 

This work primarily focused on comparison of different representations in terms of the robustness they provide. To this end, experiments were conducted on the TIMIT phoneme classification task. However, the results reported in this paper also have implications for the construction of ASR systems. In future work, we plan to investigate extensions to the proposed technique in order to facilitate the recognition of continuous speech. One straight-forward approach would be to pre-process the speech signals using the combination of the subband and cepstral SVM classifiers, and error-correcting output codes and generate class-wise feature vectors for overlapping and extended frames of speech. These feature vectors can be extracted in a manner similar to the MFCC features. An HMM-based system can then be trained using these feature vectors for recognition of continuous speech. Alternatively, the proposed technique can also be integrated with other approaches such as the hybrid phone-based HMM-SVM architecture \cite{svm2,cont2} and the token-passing algorithm \cite{svmcontnew2} for continuous speech recognition. In the former, a baseline HMM system would be required to perform a first pass through the test data and for each utterance, generate a set of possible segmentations into phonemes. The best segmentations can then re-scored by the combined SVM classifier to predict the final phoneme sequence. This approach has provided improvements in recognition performance over HMM baselines on both small and large vocabulary recognition tasks, even though the SVM classifiers were constructed solely from the cepstral representations \cite{svm2,cont2}. However, this HMM-SVM hybrid solution can also limit the efficiency of SVMs due to possible errors in the segmentation stage. On the other hand, a recognizer based solely on SVMs as discussed in \cite{svmcontnew2} can also employed which makes decisions at a frame level via SVMs and determines the chain of recognized phonemes and words using the token-passing algorithm. These extensions will be the subject of a future study.

\bibliographystyle{IEEEtran}
\bibliography{referencesfinal}

\begin{thebibliography}{10}
\providecommand{\url}[1]{#1}
\csname url@samestyle\endcsname
\providecommand{\newblock}{\relax}
\providecommand{\bibinfo}[2]{#2}
\providecommand{\BIBentrySTDinterwordspacing}{\spaceskip=0pt\relax}
\providecommand{\BIBentryALTinterwordstretchfactor}{4}
\providecommand{\BIBentryALTinterwordspacing}{\spaceskip=\fontdimen2\font plus
\BIBentryALTinterwordstretchfactor\fontdimen3\font minus
  \fontdimen4\font\relax}
\providecommand{\BIBforeignlanguage}[2]{{%
\expandafter\ifx\csname l@#1\endcsname\relax
\typeout{** WARNING: IEEEtran.bst: No hyphenation pattern has been}%
\typeout{** loaded for the language `#1'. Using the pattern for}%
\typeout{** the default language instead.}%
\else
\language=\csname l@#1\endcsname
\fi
#2}}
\providecommand{\BIBdecl}{\relax}
\BIBdecl

\bibitem{mfcc}
S.~B. Davis and P.~Mermelstein, ``{Comparison of Parametric Representations for
  Monosyllabic Word Recognition in Continuously Spoken Sentences},'' \emph{IEEE
  Trans. ASSP}, vol.~28, pp. 357--366, 1980.

\bibitem{hermansky}
H.~Hermansky, ``{Perceptual Linear Predictive ({PLP}) Analysis of Speech},''
  \emph{J. Acoust. Soc. Amer.}, vol.~87, no.~4, pp. 1738--1752, April 1990.

\bibitem{lippmann}
R.~Lippmann, ``{Speech Recognition by Machines and Humans},'' \emph{Speech
  Comm.}, vol.~22, no.~1, pp. 1--15, 1997.

\bibitem{millernicely}
G.~Miller and P.~Nicely, ``{An Analysis of Perceptual Confusions among some
  English Consonants},'' \emph{J. Acoust. Soc. Amer.}, vol.~27, no.~2, pp.
  338--352, 1955.

\bibitem{jont}
J.~B. Allen, ``{How do humans process and recognize speech?}'' \emph{IEEE
  Trans. Speech \& Audio Proc.}, vol.~2, no.~4, pp. 567--577, 1994.

\bibitem{srokabraida}
J.~Sroka and L.~Braida, ``{Human and Machine Consonant Recognition},''
  \emph{Speech Comm.}, vol.~45, no.~4, pp. 401--423, 2005.

\bibitem{meyer2007}
B.~Meyer, M.~W{\"{a}}chter, T.~Brand, and B.~Kollmeier, ``{Phoneme Confusions
  in Human and Automatic Speech Recognition},'' \emph{Proc. INTERSPEECH}, pp.
  2740--2743, 2007.

\bibitem{atal}
B.~Atal, ``{Automatic Speech Recognition: a Communication Perspective},''
  \emph{Proc. ICASSP}, pp. 457--460, 1999.

\bibitem{peters}
S.~D. Peters, P.~Stubley, and J.~Valin, ``{On the Limits of Speech Recognition
  in Noise},'' \emph{Proc. ICASSP}, pp. 365--368, 1999.

\bibitem{herve}
H.~Bourlard, H.~Hermansky, and N.~Morgan, ``{Towards Increasing Speech
  Recognition Error Rates},'' \emph{Speech Comm.}, vol.~18, no.~3, pp.
  205--231, 1996.

\bibitem{PaliwalAlsteris2003}
K.~K. Paliwal and L.~D. Alsteris, ``{On the Usefulness of STFT Phase Spectrum
  in Human Listening Tests},'' \emph{Speech Comm.}, vol.~45, no.~2, pp.
  153--170, 2005.

\bibitem{PaliwalAlsteris2006}
L.~D. Alsteris and K.~K. Paliwal, ``{Further Intelligibility Results from Human
  Listening Tests using the Short-Time Phase Spectrum},'' \emph{Speech Comm.},
  vol.~48, no.~6, pp. 727--736, 2006.

\bibitem{cmvn}
O.~Viikki and K.~Laurila, ``{Cepstral Domain Segmental Feature Vector
  Normalization for Noise Robust Speech Recognition},'' \emph{Speech Comm.},
  vol.~25, pp. 133--147, 1998.

\bibitem{mva}
C.~Chen and J.~Bilmes, ``{MVA Processing of Speech Features},'' \emph{IEEE
  Trans. ASLP}, vol.~15, no.~1, pp. 257--270, 2007.

\bibitem{vts1}
P.~J. Moreno, B.~Raj, and R.~M. Stern, ``{A Vector Taylor Series Approach for
  Environment-Independent Speech Recognition},'' \emph{Proc. ICASSP}, pp.
  733--736, 1996.

\bibitem{etsi}
E.~standard doc., ``{Speech processing, Transmission and Quality aspects (STQ):
  Advanced front-end feature extraction},'' \emph{ETSI ES 202 050}, 2002.

\bibitem{journal1}
J.~Yousafzai, Z.~Cvetkovi\'c, P.~Sollich, and B.~Yu, ``{Combined Features and
  Kernel Design for Noise Robust Phoneme Classification Using Support Vector
  Machines},'' \emph{To appear in the IEEE Trans. ASLP}, 2011.

\bibitem{fletcher}
H.~Fletcher, \emph{{Speech and Hearing in Communication}}.\hskip 1em plus 0.5em
  minus 0.4em\relax New York: Van Nostrand, 1953.

\bibitem{mcauleyming2005}
J.~McAuley, J.~Ming, D.~Stewart, and P.~Hanna, ``Subband correlation and robust
  speech recognition,'' \emph{IEEE Trans. on Speech and Audio Proc.}, vol.~13,
  no.~5, pp. 956 -- 964, 2005.

\bibitem{mingsmith2002}
J.~Ming, P.~Jancovic, and F.~Smith, ``Robust speech recognition using
  probabilistic union models,'' \emph{IEEE Trans. on Speech and Audio Proc.},
  vol.~10, no.~6, pp. 403 -- 414, Sep. 2002.

\bibitem{vaseghinew}
P.~McCourt, N.~Harte, and S.~Vaseghi, ``{Discriminative Multi-resolution
  Sub-band and Segmental Phonetic Model Combination},'' \emph{IET Electronics
  Letters}, vol.~36, no.~3, pp. 270 --271, 2000.

\bibitem{sub1}
S.~Thomas, S.~Ganapathy, and H.~Hermansky, ``Recognition {O}f {R}everberant
  {S}peech {U}sing {F}requency {D}omain {L}inear {P}rediction,'' \emph{IEEE
  Signal Process. Letters}, vol.~15, pp. 681--684, 2008.

\bibitem{sub2}
S.~Tibrewala and H.~Hermansky, ``{Subband Based Recognition Of Noisy Speech},''
  \emph{Proc. ICASSP}, pp. 1255--1258, 1997.

\bibitem{sub3}
P.~McCourt, S.~Vaseghi, and N.~Harte, ``{Multi-Resolution Cepstral Features for
  Phoneme Recognition across Speech Sub-Bands},'' \emph{Proc. ICASSP}, pp.
  557--560, 1998.

\bibitem{sub4}
H.~Bourlard and S.~Dupont, ``{Subband-based Speech Recognition},'' \emph{Proc.
  ICASSP}, pp. 1251--1254, 1997.

\bibitem{sub5}
S.~Okawa, E.~Bocchieri, and A.~Potamianos, ``{Multi-band Speech Recognition in
  Noisy Environments},'' \emph{Proc. ICASSP}, pp. 641--644, 1998.

\bibitem{sub6}
C.~Cerisara, J.-P. Haton, J.-F. Mari, and D.~Fohr, ``{A Recombination Model for
  Multi-band Speech Recognition},'' \emph{ICASSP}, pp. 717 --720 vol.2, 1998.

\bibitem{dereverb1}
J.~Flanagan, J.~Johnston, R.~Zahn, and G.~Elko, ``{Computer-Steered Microphone
  Arrays for Sound Transduction in Large Rooms},'' \emph{J. Acoust. Soc.
  Amer.}, vol.~78, no.~11, pp. 1508--1518, 1985.

\bibitem{dereverb2}
M.~Wu and D.~Wang, ``{A Two-Stage Algorithm for One-Microphone Reverberant
  Speech Enhancement},'' \emph{IEEE Trans. ASLP}, vol.~14, pp. 774--784, 2006.

\bibitem{hlmgmm}
H.~Chang and J.~Glass, ``{Hierarchical Large-Margin Gaussian Mixture Models for
  Phonetic Classification},'' \emph{Proc. ASRU}, pp. 272--275, 2007.

\bibitem{hiddencrf2}
D.~Yu, L.~Deng, and A.~Acero, ``{Hidden Conditional Random Fields with
  Distribution Constraints for Phone Classification},'' \emph{Proc.
  INTERSPEECH}, pp. 676--679, 2009.

\bibitem{shasaul}
F.~Sha and L.~K. Saul, ``{Large Margin Gaussian Mixture Modeling for Phonetic
  Classification and Recognition},'' \emph{Proc. ICASSP}, pp. 265--268, 2006.

\bibitem{rls2}
R.~Rifkin, K.~Schutte, M.~Saad, J.~Bouvrie, and J.~Glass, ``{Noise Robust
  Phonetic Classification with Linear Regularized Least Squares and
  Second-Order Features},'' \emph{Proc. ICASSP}, pp. 881--884, 2007.

\bibitem{halberstadt97}
A.~Halberstadt and J.~Glass, ``{Heterogeneous Acoustic Measurements for
  Phonetic Classification},'' \emph{Proc. EuroSpeech}, pp. 401--404, 1997.

\bibitem{clarkson}
P.~Clarkson and P.~J. Moreno, ``{On the Use of Support Vector Machines for
  Phonetic Classification},'' \emph{Proc. ICASSP}, pp. 585--588, 1999.

\bibitem{hiddencrf}
A.~Gunawardana, M.~Mahajan, A.~Acero, and J.~C. Platt, ``{Hidden Conditional
  Random Fields for Phone Classification},'' \emph{Proc. INTERSPEECH}, pp.
  1117--1120, 2005.

\bibitem{galesita}
M.~Layton and M.~Gales, ``{Augmented Statistical Models for Speech
  Recognition},'' \emph{Proc. ICASSP}, pp. I29--132, 2006.

\bibitem{classificationref1}
V.~Pitsikalis and P.~Maragos, ``{Analysis and Classification of Speech Signals
  by Generalized Fractal Dimension Features},'' \emph{Speech Comm.}, vol.~51,
  pp. 1206--1223, 2009.

\bibitem{classificationref2}
S.~Dusan, ``{On the Relevance of Some Spectral and Temporal Patterns for Vowel
  Classification},'' \emph{Speech Comm.}, vol.~49, pp. 71--82, 2007.

\bibitem{classificationref3}
K.~M. Indrebo, R.~J. Povinelli, and M.~T. Johnson, ``{Sub-banded Reconstructed
  Phase Spaces for Speech Recognition},'' \emph{Speech Comm.}, vol.~48, no.~7,
  pp. 760--774, 2006.

\bibitem{halberstadt98}
A.~Halberstadt and J.~Glass, ``{Heterogeneous Measurements and Multiple
  Classifiers for Speech Recognition},'' \emph{Proc. ICSLP}, pp. 995--998,
  1998.

\bibitem{svm2}
A.~Ganapathiraju, J.~E. Hamaker, and J.~Picone, ``{Applications of Support
  Vector Machines to Speech Recognition},'' \emph{IEEE Trans. Signal Proc.},
  vol.~52, no.~8, pp. 2348--2355, 2004.

\bibitem{cont2}
S.~E. Kr{\"u}ger, M.~Schaff{\"n}er, M.~Katz, E.~Andelic, and A.~Wendemuth,
  ``{Speech Recognition with Support Vector Machines in a Hybrid System},''
  \emph{Proc. INTERSPEECH}, pp. 993--996, 2005.

\bibitem{svmcontnew2}
J.~Padrell-Sendra, D.~Mart\'{\i}n-Iglesias, and F.~D\'{\i}az-de Mar\'{\i}a,
  ``{Support Vector Machines for Continuous Speech Recognition},'' \emph{Proc.
  EUSIPCO}, 2006.

\bibitem{vapnik}
V.~N. Vapnik, \emph{{The Nature of Statistical Learning Theory}}.\hskip 1em
  plus 0.5em minus 0.4em\relax New York: Springer-Verlag, 1995.

\bibitem{galescont}
N.~Smith and M.~Gales, ``{Speech Recognition using SVMs},'' in \emph{Adv.
  Neural Inf. Process. Syst.}, vol.~14, 2002, pp. 1197--1204.

\bibitem{sequencekernels}
J.~Louradour, K.~Daoudi, and F.~Bach, ``{Feature Space Mahalanobis Sequence
  Kernels: Application to SVM Speaker Verification},'' \emph{IEEE Trans. ASLP},
  vol.~15, no.~8, pp. 2465--2475, 2007.

\bibitem{svm1}
A.~Sloin and D.~Burshtein, ``{Support Vector Machine Training for Improved
  Hidden Markov Modeling},'' \emph{IEEE Trans. Signal Proc.}, vol.~56, no.~1,
  pp. 172--188, 2008.

\bibitem{fisher}
T.~Jaakkola and D.~Haussler, ``{Exploiting Generative Models in Discriminative
  Classifiers},'' in \emph{Adv. Neural Inf. Process. Syst.}, vol.~11, 1999, pp.
  487--493.

\bibitem{svmcontnew}
R.~Solera-Urena, D.~Mart\'{\i}n-Iglesias, A.~Gallardo-Antol\'{\i}n,
  C.~Pel\'{a}ez-Moreno, and F.~D\'{\i}az-de Mar\'{\i}a, ``{Robust ASR using
  Support Vector Machines},'' \emph{Speech Comm.}, vol.~49, no.~4, pp.
  253--267, 2007.

\bibitem{dietterichbakiri}
T.~Dietterich and G.~Bakiri, ``{Solving Multiclass Learning Problems via
  Error-Correcting Output Codes},'' \emph{J. Artif. Intell. Res.}, vol.~2, pp.
  263--286, 1995.

\bibitem{rifkin}
R.~Rifkin and A.~Klautau, ``{In Defense of One-Vs-All Classification},''
  \emph{J. Mach. Learn. Res.}, vol.~5, pp. 101--141, 2004.

\bibitem{martinv}
M.~Vetterli and J.~Kovacevic, \emph{Wavelets and {S}ubband {C}oding}.\hskip 1em
  plus 0.5em minus 0.4em\relax Englewood Cliffs, NJ: Prentice-Hall, 1995.

\bibitem{furui}
S.~Furui, ``{Speaker-Independent Isolated Word Recognition using Dynamic
  Features of Speech Spectrum},'' \emph{IEEE Trans. ASSP}, vol.~34, no.~1, pp.
  52--59, 1986.

\bibitem{dietensemble}
T.~Dietterich, ``{Ensemble Methods in Machine Learning},'' \emph{Lecture Notes
  in Computer Science: Multiple Classifier Systems}, pp. 1--15, 2000.

\bibitem{hansen}
L.~Hansen and P.~Salamon, ``{Neural Network Ensembles},'' \emph{IEEE Trans.
  PAMI}, vol.~12, no.~10, pp. 993--1001, 1990.

\bibitem{wolpert}
D.~Wolpert, ``{Stacked Generalization},'' \emph{Neural Networks}, vol.~5,
  no.~2, pp. 241--259, 1992.

\bibitem{timit}
J.~Garofolo, L.~Lamel, W.~Fisher, J.~Fiscus, D.~Pallet, and N.~Dahlgren,
  ``{TIMIT Acoustic-Phonetic Continuous Speech Corpus},'' \emph{Linguistic Data
  Consortium}, 1993.

\bibitem{kaifulee}
K.~F. Lee and H.~W. Hon, ``{Speaker-Independent Phone Recognition Using Hidden
  Markov Models},'' \emph{IEEE Trans. ASSP}, vol.~37, no.~11, pp. 1641--1648,
  1989.

\bibitem{pink1}
R.~F. Voss and J.~Clarke, ``{1/f Noise in Music: Music from 1/f Noise},''
  \emph{J. Acoust. Soc. Amer.}, vol.~63, no.~1, pp. 258--263, 1978.

\bibitem{pink2}
B.~J. West and M.~Shlesinger, ``{The Noise in Natural Phenomena},''
  \emph{American Scientist}, vol.~78, no.~1, pp. 40--45, 1990.

\bibitem{pink3}
P.~Grigolini, G.~Aquino, M.~Bologna, M.~Lukovic, and B.~J. West, ``{A Theory of
  1/f Noise in Human Cognition},'' \emph{Physica A: Stat. Mech. and its Appl.},
  vol. 388, no.~19, pp. 4192--4204, 2009.

\bibitem{icsi}
\BIBentryALTinterwordspacing
``The {ICSI} {M}eeting {R}ecorder {P}roject - {R}oom {R}esponses,'' online Web
  Resource. [Online]. Available:
  \url{http://www.icsi.berkely.edu/speech/papers/asru01-meansub-corr.html}
\BIBentrySTDinterwordspacing

\bibitem{multistyle1}
M.~Holmberg, D.~Gelbart, and W.~Hemmert, ``{Automatic Speech Recognition with
  an Adaptation Model Motivated by Auditory Processing},'' \emph{IEEE Trans.
  ASLP}, vol.~14, no.~1, pp. 43--49, 2006.

\bibitem{multistyle2}
R.~Lippmann and E.~A. Martin, ``{Multi-Style Training for Robust Isolated-Word
  Speech Recognition},'' \emph{Proc. ICASSP}, pp. 705--708, 1987.

\bibitem{gales96robust}
M.~Gales and S.~Young, ``{Robust Continuous Speech Recognition using Parallel
  Model Combination},'' \emph{IEEE Trans. SAP}, vol.~4, pp. 352--359, Sept.
  1996.

\bibitem{isit}
J.~Yousafzai, Z.~Cvetkovi\'c, and P.~Sollich, ``{Towards Robust Phoneme
  Classification with Hybrid Features },'' \emph{Proc. ISIT}, pp. 1643--1647,
  2010.

\bibitem{malah}
Y.~Ephraim and D.~Malah, ``{Speech Enhancement Using a Minimum Mean-Square
  Error Short-time Spectral Amplitude Estimator},'' \emph{IEEE Trans. ASSP},
  vol. ASSP-32, pp. 1109--1121, 1984.

\bibitem{malah2}
------, ``{Speech Enhancement Using a Minimum Mean-Square Log-Spectral
  Amplitude Estimator},'' \emph{IEEE Trans. ASSP}, vol. ASSP-33, pp. 443--445,
  1985.

\end{thebibliography}

\end{document}